\documentclass[acmsmall,screen,nonacm]{acmart}

\setcopyright{none}
\acmYear{2026}
\acmVolume{}
\acmNumber{}
\acmArticle{}
\acmMonth{9}
\acmDOI{}
\acmPrice{}
\microtypesetup{expansion=false} 
\usepackage{graphicx}
\usepackage{booktabs}
\usepackage{tabularx}
\usepackage{longtable}
\usepackage{array}
\usepackage{ragged2e}
\usepackage{amsmath}
\usepackage{enumitem}
\usepackage{xcolor}
\usepackage{colortbl}
\usepackage{xurl}
\usepackage{tikz}
\usepackage{placeins}
\usetikzlibrary{positioning,arrows.meta,backgrounds,fit,calc,shapes.geometric,shapes.symbols,shapes.arrows}

\hypersetup{hidelinks,pdfsubject={Empirical study of long-horizon LLM agent failures}}
\setlist[itemize]{leftmargin=1.35em,topsep=0.15em,itemsep=0.2em}
\setlist[enumerate]{leftmargin=1.55em,topsep=0.15em,itemsep=0.2em}
\renewcommand{\arraystretch}{1.08}
\newcolumntype{P}[1]{>{\RaggedRight\arraybackslash}p{#1}}

\title[Dependable Long-Horizon LLM Agents]{Dependable Long-Horizon {LLM} Agents: A Trajectory-Centered Empirical Study of Where Errors Enter and Which Controls Reach Them}

\author{Shaohuang Wang}
\affiliation{\institution{Villanova University}\department{Department of Electrical and Computer Engineering}\city{Villanova}\state{Pennsylvania}\country{USA}}
\email{swang08@villanova.edu}
\author{Jinghao Wen}
\affiliation{\institution{Villanova University}\department{Department of Electrical and Computer Engineering}\city{Villanova}\state{Pennsylvania}\country{USA}}
\email{jwen01@villanova.edu}
\author{Xulei Zhang}
\affiliation{\institution{Villanova University}\department{Department of Electrical and Computer Engineering}\city{Villanova}\state{Pennsylvania}\country{USA}}
\email{xzhang05@villanova.edu}
\author{Xun Jiao}
\affiliation{\institution{Villanova University}\department{Department of Electrical and Computer Engineering}\city{Villanova}\state{Pennsylvania}\country{USA}}
\email{xun.jiao@villanova.edu}
\renewcommand{\shortauthors}{Wang et al.}

\newcommand{\nNann}{96}
\newcommand{\nNfail}{95}

\newcommand{\nNmad}{8}
\newcommand{\nNhc}{29}
\newcommand{\nNag}{59}
\newcommand{\nEvYes}{68}
\newcommand{\nEvYesPct}{72}

\newcommand{\nEvUnusedPct}{88}
\newcommand{\nEvUsed}{7}
\newcommand{\nEvAck}{19}
\newcommand{\nEvSeen}{41}
\newcommand{\nEvBefore}{32}
\newcommand{\nDet}{29}
\newcommand{\nDetPct}{31}
\newcommand{\nResp}{34}

\newcommand{\nRespPartial}{4}
\newcommand{\nDistMed}{5.5}
\newcommand{\nDistMean}{17.9}
\newcommand{\nDistMax}{126}
\newcommand{\nOriginPosMed}{0.18}
\newcommand{\nDistRelMed}{0.67}
\newcommand{\nGoldN}{88}
\newcommand{\nGoldSame}{40}
\newcommand{\nGoldSamePct}{45}
\newcommand{\nGoldWOne}{52}
\newcommand{\nGoldWOnePct}{59}
\newcommand{\nGoldEarlier}{37}
\newcommand{\nGoldLater}{11}
\newcommand{\nGoldGapMed}{3}

\newcommand{\nGoldHcWOnePct}{45}

\newcommand{\nGoldAgWOnePct}{66}

\newcommand{\nExtNone}{92}
\newcommand{\nRestoreSuff}{35}
\newcommand{\nRestorePlan}{56}
\newcommand{\nSecondary}{34}
\newcommand{\nConfHigh}{38}
\newcommand{\nConfMed}{55}
\newcommand{\nConfLow}{2}
\newcommand{\nWWhcN}{58}
\newcommand{\nWWhcCrash}{26}
\newcommand{\nWWhcFallback}{25}
\newcommand{\nWWhcCrashPct}{45}
\newcommand{\nWWagCrash}{0}
\newcommand{\nWWagN}{126}
\newcommand{\nFallbackBlameOrch}{4}
\newcommand{\nFallbackBlameOther}{21}

\newcommand{\nHopsMed}{5}
\newcommand{\nHopsMin}{2}
\newcommand{\nHopsMax}{8}
\newcommand{\nDomOne}{10}
\newcommand{\nDomTwo}{42}
\newcommand{\nDomThreePlus}{43}
\newcommand{\nSecIndep}{26}
\newcommand{\nSecHarness}{8}
\newcommand{\nDistDecision}{6}
\newcommand{\nCountDecision}{35}
\newcommand{\nDistFabrication}{5}
\newcommand{\nCountFabrication}{25}
\newcommand{\nDistIntegration}{8.5}
\newcommand{\nCountIntegration}{10}

\newcommand{\nNoOutput}{23}
\newcommand{\nWrongOutput}{72}
\newcommand{\nModesPerRun}{2.9}
\newcommand{\nMastIncVerKinds}{5}
\begin{document}
\begin{abstract}
When an LLM agent fails part-way through a long task, the usual analysis records a symptom: a failure-mode label, or the agent and step judged responsible. It does not say where the error entered, which state carried it to the symptom, whether anything in the trajectory contradicted it, or which control could have caught it. We define an episode record that ties a candidate fault to its origin, propagation, detection, response and consequence, located by operational phase and state domain, and that names the control families whose observation and reach cover it. We turned the record into a written annotation protocol and applied it to \nNfail{} failed runs of six multi-agent systems from the public Who\&When and MAD corpora, reading each trajectory in full. Faults enter early (median at \nOriginPosMed{} of the run), mostly as wrong decisions or fabrications over task state, and in \nGoldEarlier{} of \nGoldN{} Who\&When runs before the step the corpus' human labels mark as decisive. In \nEvYesPct\% of failed runs the trajectory already contained evidence against the error before the failure; in \nEvUnusedPct\% of those no agent acted on it. Something flagged a deviation in \nDetPct\% of runs, but no response removed the deviated state. Independent judges reading the unused evidence, task contracts fixed at admission, and re-checking beliefs against the environment reach most episodes; restoring the agent's own state would have sufficed in \nRestoreSuff{} of \nNfail{} runs. A text scan of the raw corpus also shows that \nWWhcCrashPct\% of the hand-crafted Who\&When failures ended with a provider content-filter crash after which the harness emitted a final answer no agent produced, a failure class that neither symptom taxonomies nor responsible-agent labels represent. Protocol, records and scripts are included.
\end{abstract}

\maketitle
\section{Introduction}
\label{sec:intro}

Large language model (LLM) agents carry out tasks by selecting tools, observing the results, and revising their actions over many steps~\citep{yao2023react,yang2024sweagent,wang2024openhands,packer2023memgpt}. For such systems, dependability is a property of the whole \emph{trajectory}: the sequence of decisions and state changes from the first step to the last. Long trajectories change how errors behave. A retrieved statement can introduce a false assumption. Memory can keep that assumption alive, another agent can reuse it, and an action can turn it into a change in an external system that no checkpoint can undo. In a simulated vending business, every evaluated model had runs that derailed through misread delivery schedules, forgotten orders, or repeated loops~\citep{backlund2025vendingbench}. In 1,794 annotated coding-agent trajectories, the first error typically occurred early, while its symptoms appeared only much later~\citep{zhao2026failureprocess}. In current checkpoint-and-rollback frameworks, a restored agent can re-issue actions that were already committed, including payments~\citep{wu2026saferesume,zheng2026acrfence}.

One released trajectory shows how these effects combine~\citep{xu2024theagentcompany}. An agent in a simulated software company is asked to obtain a repository location from a colleague, Alex Turner, through the company chat, then clone the repository and start a server. Logged into the chat as an administrator, it opens the user administration page and renames the account it is logged in with to ``Alex Turner''. It then sends its question to a different colleague, addressed to Alex, and waits. For the remaining 81 turns it reports that it is still waiting for a reply, until the run ends 22 minutes later with nothing cloned and the renamed account never restored. No single step is a familiar model error. The damage comes from how the steps connect: an action outside the task's scope, a message to the wrong recipient, a belief that the message reached its target, and no point at which the run stops to check. This trajectory serves as a running example in Section~\ref{sec:execution}.

Classical dependable computing gives us the vocabulary and tools for such problems. It distinguishes faults, errors, and service failures, and it offers mature theories of checkpointing, compensation, and redundancy~\citep{avizienis2004taxonomy}. LLM agents, however, give three of its classical problems new forms. (1)~\emph{Requests are generated.} Conventional systems make retries safe with idempotence keys and deduplication, but an agent's model re-generates the request, so a retry may differ in content and form, and a key derived from the original request may not match it. (2)~\emph{State is partly semantic.} Integrity checks confirm that a stored item is intact; in an agent, an intact item can still be wrong, because its correctness depends on where it came from and what it means. (3)~\emph{Effects reach services the agent does not control.} As in distributed transactions, restoring local state leaves committed external effects in place, and agent frameworks often do not record which effects were already released.

When a run like this is analyzed, the analysis usually stops at the symptom. The failure gets a mode label such as ``task derailment'' or ``incorrect verification''~\citep{cemri2025mast}, or an annotator names the agent and step that were responsible~\citep{zhang2025whowhen}. Both are useful, and both leave out what one would need to fix the system: where the error actually entered, which part of the agent's state carried it to the point where it became visible, whether anything in the trajectory contradicted it along the way, and which kind of control could have caught it. As far as we know, nobody has measured these things on released trajectories at any scale.

That is what this paper does. We define an \emph{episode record} that ties a candidate fault to its origin, its propagation, its detection and response, and its consequence. Each event is located by operational phase (P0--P7) and state domain (R1--R6), and the record names the control families whose observation and reach would cover the episode (Section~\ref{sec:execution}). We wrote the record down as an annotation protocol and applied it to failed trajectories from two public corpora, Who\&When~\citep{zhang2025whowhen} and the MAST/MAD traces~\citep{cemri2025mast}, reading each trajectory in full (Section~\ref{sec:study}). Three questions:
\begin{itemize}
\item \textbf{RQ1.} In which phase and state domain do faults enter, how far before the visible symptom, and how does the origin relate to the ``decisive step'' that the corpus authors labelled?
\item \textbf{RQ2.} When a run fails, was there evidence in the trajectory that contradicted the error? Did any agent use it? Did detection, when it happened, lead to a response that worked?
\item \textbf{RQ3.} Which control families would have observed and reached each episode, and how often would restoring the agent's own state have been enough?
\end{itemize}

We completed \nNfail{} failed runs from six multi-agent systems (Section~\ref{sec:results}). The short version of what we found:
\begin{itemize}
\item Faults enter early and are mostly decisions over task state. Origins cluster in deliberation and execution (P1--P2) over task and knowledge state (R2), at a median of \nOriginPosMed{} of the run; the median gap to the symptom is \nDistRelMed{} of the run. In \nGoldEarlier{} of \nGoldN{} Who\&When traces our origin is earlier than the step the corpus marks as decisive.
\item The evidence is usually there and usually ignored. \nEvYesPct\% of failed runs contain, before the failure, an observation or message that contradicts the erroneous belief; in \nEvUnusedPct\% of those nobody acted on it. Something flagged a deviation in \nDetPct\% of runs. Not one response removed the deviated state.
\item The controls that would reach these episodes are not the ones people mostly build. Independent judges that read the unused evidence, task contracts, provenance on stored facts, and re-checking beliefs against the environment cover most episodes. A local checkpoint alone would have helped in \nRestoreSuff{} runs; in \nRestorePlan{} the fault sat in a plan or a policy that a restore would simply replay.
\item A harness-level failure hides behind the attribution labels. In \nWWhcCrash{} of the \nWWhcN{} hand-crafted Who\&When runs (\nWWhcCrashPct\%) a provider content filter killed the orchestrator and the harness emitted a final answer that no agent had produced. The corpus attributes \nFallbackBlameOther{} of these \nWWhcFallback{} runs to an agent other than the orchestrator.
\end{itemize}
The protocol, all annotation records and the scripts that produce every number in the paper are included as ancillary files. Appendix~\ref{sec:background} gives the background the framework draws on; readers who want the dependability vocabulary and the control families in detail should start there.

\section{A Framework for Trajectory Dependability}
\label{sec:execution}

\begin{figure}[t]
\centering\resizebox{.98\linewidth}{!}{\definecolor{smA}{HTML}{7040D8}\definecolor{smB}{HTML}{1766B1}\definecolor{smC}{HTML}{078557}\definecolor{smD}{HTML}{D54832}\definecolor{smInk}{HTML}{20242A}\definecolor{smG}{HTML}{6B6F76}
\begin{tikzpicture}[x=1cm,y=1cm,font=\sffamily\small,
 ph/.style={draw=smB,fill=smB!5,rounded corners=3pt,line width=.6pt,align=center,minimum height=1.05cm,text width=2.0cm,text=smB!80!black},
 pg/.style={draw=smG,fill=smG!5,rounded corners=3pt,line width=.6pt,align=center,minimum height=1.05cm,text width=2.3cm,text=smInk},
 rd/.style={draw=smC,fill=smC!5,rounded corners=3pt,line width=.6pt,align=center,minimum height=1.0cm,text width=4.45cm,text=smC!70!black,font=\sffamily\footnotesize},
 hd/.style={draw=smA,fill=smA!5,rounded corners=3pt,line width=.6pt,anchor=west,minimum height=.5cm,text width=9.3cm,text=smInk},
 ar/.style={-{Stealth[length=4pt]},line width=.6pt,draw=smInk!70},
 ttl/.style={font=\sffamily\normalsize\bfseries,anchor=west,text=white,rounded corners=3pt,inner xsep=6pt,inner ysep=3pt,minimum width=16.45cm,align=left}]
\node[ttl,fill=smB] at (-1.3,6.55) {1\quad Operational phases (P0--P7): when an operation occurs};
\node[pg] (p0) at (0,4.6) {\textbf{P0}\\Scope, authorize};
\node[ph] (p1) at (2.65,4.6) {\textbf{P1}\\Deliberate};
\node[ph] (p2) at (5.05,4.6) {\textbf{P2}\\Dispatch, execute};
\node[ph] (p3) at (7.45,4.6) {\textbf{P3}\\Observe, integrate};
\node[ph] (p4) at (9.85,4.6) {\textbf{P4}\\Verify, decide};
\node[ph] (p5) at (7.45,3.05) {\textbf{P5}\\Recover, escalate};
\node[pg] (p6) at (14.0,4.6) {\textbf{P6}\\Terminate, hand off};
\node[pg] (p7) at (14.0,3.05) {\textbf{P7}\\Audit, govern\\(task-wide)};
\draw[smB,dashed,rounded corners=5pt,line width=.6pt] (1.45,2.3) rectangle (11.05,6.05);
\draw[ar] (p0)--(p1); \draw[ar] (p1)--(p2); \draw[ar] (p2)--(p3); \draw[ar] (p3)--(p4);
\draw[ar] (p4)--node[above=1pt,smInk]{if complete} (p6);
\draw[ar,draw=smB] (p4.north) -- ++(0,.5) -| (p1.north) node[pos=.25,above,smB,font=\sffamily\footnotesize]{repeat as needed};
\draw[ar,draw=smB] (p4.south) |- (p5.east) node[pos=.25,right,smB]{if needed};
\draw[ar,draw=smB] (p5.west) -| (p1.south) node[pos=.3,below,smB]{if safe to resume};
\node[ttl,fill=smC] at (-1.3,1.75) {2\quad State domains (R1--R6): which state an error affects};
\node[rd] at (0.95,0.75) {\textbf{R1 Model and compute}\\hardware, weights, KV storage};
\node[rd] at (5.95,0.75) {\textbf{R2 Task and knowledge}\\goals, plans, context, memory};
\node[rd] at (10.95,0.75) {\textbf{R3 Action/observation interface}\\schemas, parsers, browser, shell};
\node[rd] at (0.95,-0.5) {\textbf{R4 Environment and artifacts}\\files, databases, released effects};
\node[rd] at (5.95,-0.5) {\textbf{R5 Coordination state}\\shared memory, messages, handoffs};
\node[rd] at (10.95,-0.5) {\textbf{R6 Human and organizational}\\approvals, policy, accountability};
\node[ttl,fill=smA] at (-1.3,-1.65) {3\quad Error dynamics (H1--H7): how an error persists or spreads};
\foreach \i/\n/\d in {1/Repeated exposure/more opportunities for an event, 2/Contamination/a deviated state is retained and reused, 3/Drift/divergence from an explicit reference, 4/Cascade/one error influences several later events, 5/External impact/restoration omits released consequences, 6/Observability debt/diagnosis needs missing dependencies, 7/Resource exhaustion/execution or recovery exceeds budget}
  \node[hd] (h\i) at (-1.25,{-1.85-0.62*\i}) {\textbf{\textcolor{smA!80!black}{H\i}\ \ \n}\,---\,\d};
\foreach \i in {1,...,7} \draw[smA,line width=.6pt] (h\i.east) -- (8.8,{-1.85-0.62*\i});
\draw[smA,line width=.6pt] (8.8,-2.47) -- (8.8,-6.19);
\node[draw=smD,fill=smD!5,rounded corners=3pt,align=center,text width=3.4cm,line width=.6pt,text=smD!75!black] (ctl) at (12.0,-2.75) {\textbf{Controls}\\inspect, restrict,\\restore, or escalate};
\node[draw=smG,fill=smG!5,rounded corners=3pt,align=center,text width=3.4cm,line width=.6pt,text=smInk] (out) at (12.0,-5.35) {\textbf{Possible outcomes}\\task failure\\policy violation\\unrecovered impact};
\draw[ar,draw=smD] (ctl) -- node[right,smD!75!black,font=\sffamily\footnotesize]{may interrupt} (out);
\draw[ar,draw=smA] (8.8,-5.35) -- (out.west);
\end{tikzpicture}}
\caption{The trajectory-centered dependability framework. P0--P7 locate operations in execution; R1--R6 locate the state or resource affected; H1--H7 describe possible dynamics across these coordinates. Recovery is conditional and audit can span the task. The lower panel is an analytical synthesis: controls can interrupt the paths, and no single dynamic implies failure.}
\Description{Three layers show an execution loop with conditional recovery and cross-cutting audit, six state domains, and seven non-sequential error dynamics. Controls may limit task failure, policy violation, and unrecovered impact.}
\label{fig:framework}
\end{figure}
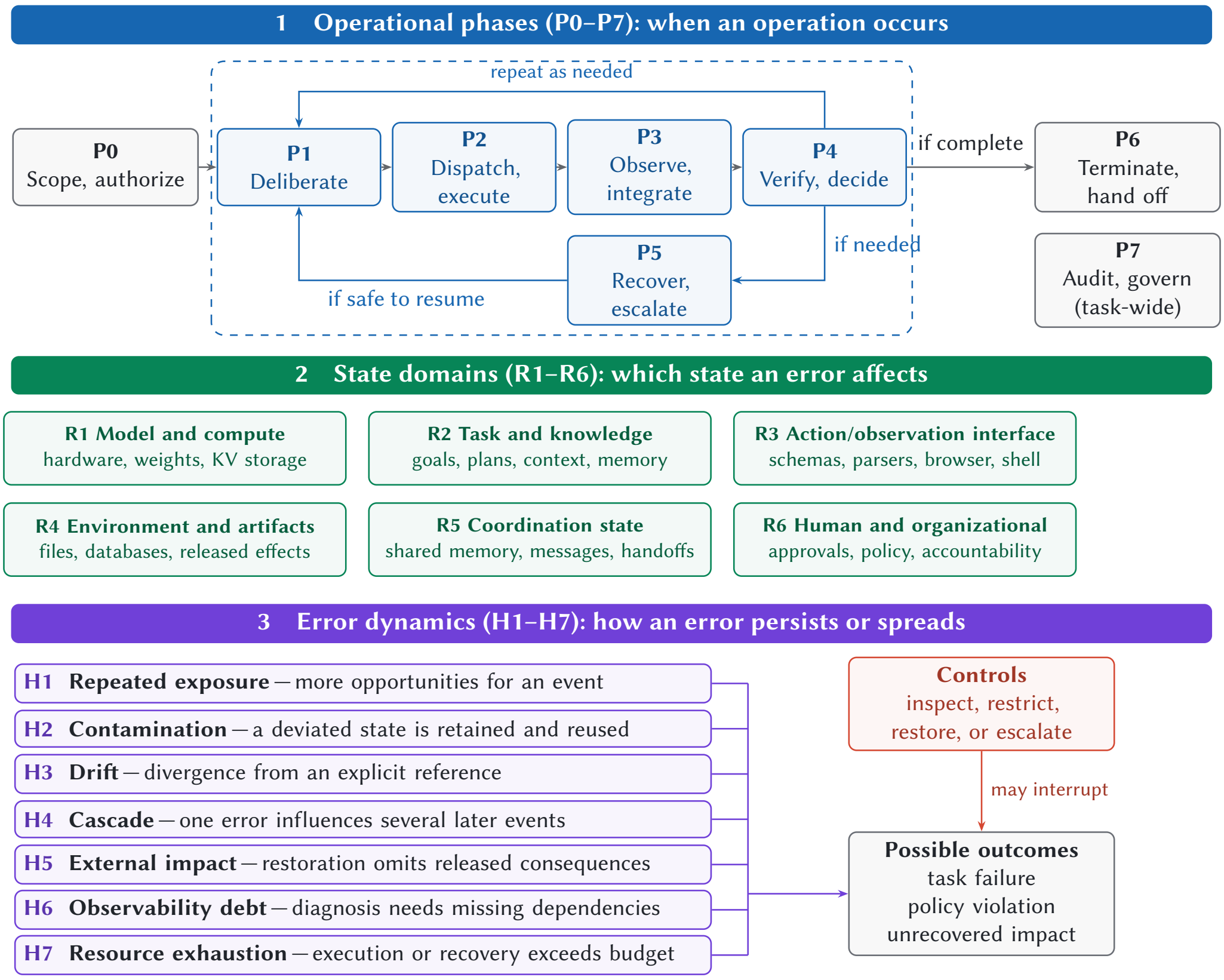

Figure~\ref{fig:framework} connects execution phases, affected state domains, and the dynamics through which errors persist or spread; this section defines each element.

\subsection{Service properties and the unit of execution}
\label{sec:classical}

An agent is a model plus a \emph{harness} that builds context, calls tools, tracks state, and schedules checks and recovery. Dependability is delivering service that can justifiably be trusted under stated conditions~\citep{avizienis2004taxonomy,laprie1985dependability}. Its attributes (availability, reliability, safety, integrity, maintainability) need separate measures because they can conflict. Stopping, for example, can preserve safety but block completion. This study covers the attributes a single run can violate: reliability (task completion), safety (policy-compliant completion or safe termination), and integrity (the task's invariants hold, and injected content gains no authority over actions); availability and maintainability enter only through oversight and audit (Sections~\ref{sec:ch6} and~\ref{sec:mech6}). We use \emph{dependability} for the whole and \emph{reliability} for the single attribute, and read the broader use of reliability in agent papers as dependability.

Let a task specification be $\mathcal{S}=(G,\mathcal{I},\mathcal{A},B)$, with outcome predicate $G$, required invariants $\mathcal{I}$, authorized actions $\mathcal{A}$, and resource budget $B$. A trajectory $\tau$ is the alternating sequence of states, actions, and observations $x_0,a_0,o_1,x_1,\ldots,a_{T-1},o_T,x_T$, where $x_t$ is the relevant, partly observable agent and environment state, $a_t$ an action, and $o_{t+1}$ its observed result; the dynamics may be nonstationary and non-Markov. In the running example from Section~\ref{sec:intro}, $\tau$ is the released log of 201 events, $G$ requires a cloned repository and a running server, and $\mathcal{A}$ does not include writes to the user directory.

\emph{Task completion} means $G(x_T)$ holds. \emph{Policy-compliant completion} also requires $a_t\in\mathcal{A}$ for all $t$, with each invariant in $\mathcal{I}$ holding where applicable. \emph{Safe termination} instead permits a specified abstention or escalation.

A \emph{fault} is a candidate cause of an error, an \emph{error} a deviation in state, and a \emph{failure} a violation of the service specification. The distinction matters because a malformed call can be a recovered task error, but an unauthorized disclosure is a safety failure whatever the final answer.

\subsection{Measuring the horizon}
\label{sec:traj}

The realized horizon of a run has several components: loop iterations, tool calls, tokens, wall-clock time, external mutations, participating agents, and sessions. The running example is long in turns (98 agent actions) and time (22 minutes) but contains only two external mutations, a renamed account and a message sent to the wrong person, and these are what matter. No single step count defines a long task, so we call a task \emph{long-horizon} when its outcome materially depends on retaining state, maintaining dependencies, or recovering across repeated interactions. Benchmarks stress different components: Vending-Bench runs exceed 20 million tokens with simple decisions~\citep{backlund2025vendingbench}, whereas RE-Bench fixes wall-clock budgets of up to eight hours~\citep{wijk2024rebench}. The \emph{required} horizon also differs from the \emph{realized} horizon, because repeated failures can turn a twenty-operation task into eighty, so observed length often reflects error. Human-equivalent task duration instead measures how demanding a task is, and its 50\% completion horizon has doubled roughly every seven months since 2019~\citep{kwa2025timehorizon}.

\subsection{Operational phases and state domains}
\label{sec:phases}
\label{sec:domains}

The framework separates \emph{when an operation occurs} from \emph{which state it affects} (Table~\ref{tab:coordinates}). P1--P5 describe recurring operations, whereas P0, P6, and P7 describe admission, delivery, and governance. These labels describe roles, not execution order.

\begin{table}[t]
\centering\small
\caption{Two complementary coordinates. A physical object can support several semantic roles; classify by the role relevant to the claim and retain secondary domains. A commit may occur during P2, not only at task termination P6.}
\label{tab:coordinates}
\begin{tabularx}{\linewidth}{@{}P{.07\linewidth}P{.27\linewidth}X@{}}
\toprule\textbf{Code} & \textbf{Coordinate} & \textbf{Operational definition}\\\midrule
P0 & Scope and authorize & Establish goal, constraints, acceptance criteria, permissions, and budget.\\
P1 & Deliberate and decompose & Select or revise a plan, action, or delegation.\\
P2 & Dispatch and execute & Issue and execute a tool, model, agent, or human operation.\\
P3 & Observe and integrate & Interpret a result and update working or persistent state.\\
P4 & Verify and decide & Assess validity, fidelity, progress, or completion.\\
P5 & Recover and escalate & Retry, replan, restore, compensate, or transfer control.\\
P6 & Terminate and hand off & Deliver or accept completion, publish a result, or transfer responsibility.\\
P7 & Audit and govern & Review traces and incidents, assign ownership, and update policy.\\\midrule
R1 & Model and compute & Hardware, kernels, weights, decoding, scheduling, physical KV storage.\\
R2 & Task and knowledge state & Goals, plans, context, evidence, summaries, and individual memory.\\
R3 & Action/observation interface & Schemas, parsers, API clients, browser and shell processes, connectors.\\
R4 & Environment and artifacts & Repositories, files, databases, services, content, and external effects.\\
R5 & Coordination state & Shared memory, messages, ownership, delegation, and handoff records.\\
R6 & Human/organizational control & Approvals, policies, audit access, accountability, and incident response.\\
\bottomrule
\end{tabularx}

\end{table}

\subsection{Episodes, causal links, and uncertainty}
\label{sec:episode}

An \emph{episode} is an analytical record: a chain or branching subgraph of the trajectory that links a candidate fault to the states, responses, and consequences it affects:
\begin{equation}
E=\langle O,\mathcal{I}_E,\mathcal{L},D,C,M,\sigma,U\rangle.
\label{eq:episode}
\end{equation}
The record captures the origin $O$, affected invariants $\mathcal{I}_E$, dependency links $\mathcal{L}$, detections $D$, responses $C$, manifestations $M$, evaluation scale $\sigma$, and uncertainty $U$. Each event retains its phase, domain, order, version, and source; H1--H7 label the links. Each dependency is marked as observed, intervention-supported, or hypothesized.

In the running example, the origin $O$ is the deliberation step (P1) that chose to rename an account rather than search the directory; the affected invariant is the authorization scope $\mathcal{A}$. The action changes an identity record in a shared system (R4, R6). The link $\mathcal{L}$ runs from the message sent to the wrong colleague (R5) to the belief (R2) that Alex Turner has been asked, which is contaminated (H2). Two observations displayed the evidence, but $D$ and $C$ are empty, and the manifestation $M$ is 81 turns of waiting (H7). Restoring the agent's own state would not undo the renamed account, which is why the recovery boundary is part of the record. Section~\ref{sec:study} turns this record into an annotation protocol and applies it at scale.

\section{Study Design}
\label{sec:study}

\subsection{Corpora}
\label{sec:corpora}

We wanted failed runs of real multi-agent systems with the complete trajectory released, not a summary of it. Two public corpora fit.

\paragraph{Who\&When}~\citep{zhang2025whowhen} contains 184 failed runs on GAIA and AssistantBench questions. 58 are \emph{hand-crafted} runs of Magentic-One (an orchestrator with a web surfer, a file surfer, a coder and a terminal agent); 126 are \emph{algorithm-generated} runs of AG2 CaptainAgent teams, where a manager assembles a group of role-named experts. Every run has a human label that names the responsible agent and the \emph{decisive mistake step}. We did not use these labels during annotation; they serve as an outside reference afterwards (Section~\ref{sec:validity}).

\paragraph{MAD}~\citep{cemri2025mast}, human-labelled split. 19 traces from AG2, MetaGPT, ChatDev, HyperAgent, AppWorld, Magentic-One and OpenManus, each labelled by three annotators with failure modes from a development version of the MAST taxonomy. We keep the majority labels for reference. The traces are raw logs in several formats and some are very long.

Each trajectory was normalized into one text file. Who\&When messages become numbered steps (\texttt{\#\#\# step $i$ [role/name]}); MAD logs are kept as they are, with line numbers. We dropped seven traces longer than 200{,}000 characters (one from Who\&When, six from MAD) because nobody, human or model, reads those in one pass.

\subsection{Annotation protocol}
\label{sec:protocol}

The protocol is a written document of about four pages; it is included with the submission. It turns the episode record of Equation~\eqref{eq:episode} into 38 fields with fixed vocabularies. For each failed run the annotator records one \emph{primary episode}: the earliest candidate fault whose consequences reach the final failure. A second, causally independent fault can be recorded as a \emph{secondary} episode. The fields that matter most for the results:

\begin{itemize}
\item \textbf{Origin.} Step, phase (P0--P7), domain (R1--R6), and kind. The kind follows a short decision tree, because the pilot showed that annotators otherwise disagree on it: \emph{specification} if the brief itself was defective at admission and the agent could not have known; \emph{fabrication} if an unobserved fact, result or placeholder argument was asserted; \emph{integration} if an observation that was in the agent's input was misread or dropped; \emph{decision} if the choice was wrong although the information was adequate; and \emph{interface}, \emph{environment}, \emph{coordination} for faults in tools, external systems and handoffs.
\item \textbf{Propagation.} The hops at which the deviated state was reused, the domains crossed, and which of the dynamics H1--H7 (Table~\ref{tab:dynamics}) apply.
\item \textbf{Symptom.} The step at which the failing output is emitted as final, or at which the run ends, and the distance from the origin.
\item \textbf{Evidence.} Whether some observation or message before the symptom contradicted the erroneous belief, and what happened to it: \emph{used}, \emph{acknowledged} but not acted on, in an agent's input but \emph{never mentioned}, or never seen by any agent.
\item \textbf{Detection and response.} Who, if anyone, flagged a deviation (the agent itself, another agent, a tool or verifier), what response followed (retry, replan, rollback, compensation, escalation, termination), and whether the response actually removed the deviated state.
\item \textbf{Controls.} Every control family from Table~\ref{tab:controlcontracts} whose inspection basis would have seen the deviation and whose boundary reaches the affected state, with the most direct one marked as primary. A separate field, \emph{restore reach}, asks whether restoring the agent's own state to just before the origin would have been enough.
\item \textbf{Harness fault.} Whether a model-provider error ended the loop, and whether the harness itself emitted a final answer that no agent step produced. We added this field after the pilot, for reasons that Section~\ref{sec:res-harness} makes clear.
\end{itemize}
Every step-valued field has to point at a concrete step or line of the normalized trace, so a reader can check it.

\paragraph{Pilot.} We first applied version 0.1 of the protocol to 20 traces chosen to span the length distribution. The annotator reports came back with 14 points where the vocabulary did not fit what was in the trace: runs whose final answer was correct but whose process failed, evidence that was present before the fault was even made, provider crashes that coincided with a fabricated answer, and so on. Version 0.2 resolves these, mostly by adding a value or a rule. The 20 pilot records were migrated to 0.2.

\subsection{Who annotated, and how many}
\label{sec:procedure}

The annotation was done by an LLM (Claude, Anthropic), one independent instance per trace, instructed with the protocol, the field schema and one finished example record. Each instance read the whole normalized trajectory before writing anything. Records were checked mechanically against the schema and the controlled vocabularies; none failed. The trajectories were treated as data throughout, never as instructions to the annotator.

We did not get through the whole corpus. Within the budget of this study we completed \nNann{} records: \nNhc{} hand-crafted and \nNag{} algorithm-generated Who\&When runs and \nNmad{} MAD traces. The sample is the 20 pilot traces plus the first annotation batches; batches were formed by balancing character counts, so there is no selection on content. For Who\&When the sample matches the corpus in run length (median 33 vs.\ 32.5 steps for hand-crafted runs, 9 vs.\ 10 for algorithm-generated ones). The MAD part is small and skewed toward short traces by the length cap, and we treat it accordingly. One MAD run turned out to have a correct final answer and a purely procedural failure; it is kept in the records but left out of the outcome statistics, which leaves \nNfail{} failed runs. The annotator rated its own confidence high for \nConfHigh{} records, medium for \nConfMed{} and low for \nConfLow{}.

\subsection{Checks that do not depend on the annotator}
\label{sec:validity}

Two. First, for all \nGoldN{} annotated Who\&When runs we compare our origin step with the human-labelled decisive step. The two should not be expected to coincide exactly, since they answer different questions: the decisive step is the step whose correction would have avoided the failure, our origin is the earliest step at which the state deviated in a way that led to it. An origin at or before the decisive step is consistent with both readings; an origin after it is a real disagreement, and we look at those cases. Second, the harness-fault result in Section~\ref{sec:res-harness} comes from a deterministic text scan over all 184 raw Who\&When trajectories and does not involve any annotation.

\section{Results}
\label{sec:results}

Everything below is computed by the included scripts from the \nNfail{} failed-run records, except where we say otherwise. Percentages are of those runs.

\subsection{RQ1: Where faults enter and how far they travel}
\label{sec:res-origin}

\paragraph{Phase and domain.} Table~\ref{tab:origin} cross-tabulates where the primary episodes start. Most faults enter while the agent executes (P2, 34 runs) or deliberates (P1, 24), and the first thing they corrupt is almost always task and knowledge state (R2, 71 runs): a plan, a fact sheet, a belief about what a page showed. Admission-time faults (P0, 13 runs) are mostly coordination-state faults (R5, 10 runs, all in CaptainAgent teams): the manager handed its group a garbled or substituted task description, so the real question never reached the agents that answered it. One origin is in model or compute state (R1), two are in the external environment (R4), and none is in human or organizational state (R6); the corpora simply contain no humans.

\begin{table}[t]
\centering\small
\caption{Origin of the primary episode by operational phase and state domain ($n=\nNfail$ failed runs).}
\label{tab:origin}
\begin{tabular}{@{}lrrrrrrr@{}}
\toprule
Phase&R1&R2&R3&R4&R5&R6&Total\\\midrule
P0&--&2&1&--&10&--&13\\
P1&--&24&--&--&--&--&24\\
P2&--&26&6&2&--&--&34\\
P3&--&9&1&--&--&--&10\\
P4&1&5&2&--&1&--&9\\
P5&--&4&--&--&--&--&4\\
P6&--&1&--&--&--&--&1\\
\midrule Total&1&71&10&2&11&0&95\\
\bottomrule
\end{tabular}
\end{table}

\paragraph{Kind.} Wrong decisions made with adequate information are the largest class (42\%, Table~\ref{tab:kind}). \emph{Fabrication} is second (26\%): an agent asserted a fact, a lookup result, or a verification it never performed, or passed a placeholder argument (for example \texttt{YOUR\_GOOGLE\_MAPS\_API\_KEY} or a path named \texttt{path\_to\_local\_copy}) to a tool that then reported success. Misread or dropped observations make up another 11\%.

\begin{table}[t]
\centering\small
\caption{Kind of the origin event.}
\label{tab:kind}
\begin{tabular}{@{}lrr@{}}
\toprule Kind&Runs&Share\\\midrule
decision&40&42\%\\
fabrication&25&26\%\\
integration&10&11\%\\
coordination&8&8\%\\
specification&4&4\%\\
interface&7&7\%\\
environment&1&1\%\\
\bottomrule
\end{tabular}
\end{table}

\begin{figure}[t]
\centering\includegraphics[width=\linewidth]{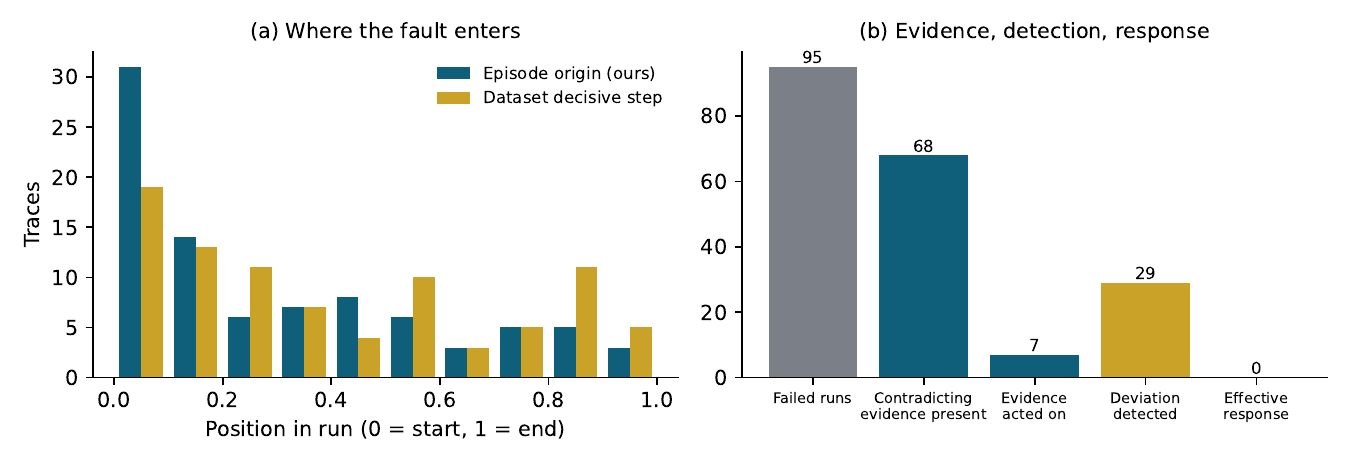}
\caption{(a) Relative position of the episode origin (ours) and of the dataset's human-labelled decisive step in \nGoldN{} Who\&When runs. (b) Of \nNfail{} failed runs: runs with contradicting evidence before the symptom, runs in which an agent acted on it, runs in which a deviation was flagged, and runs in which the response removed the deviated state.}
\label{fig:results}
\end{figure}

\paragraph{Onset and symptom.} On Who\&When, the origin lies at a median of \nOriginPosMed{} of the run length, and the median distance to the symptom is \nDistMed{} steps (mean \nDistMean{}, maximum \nDistMax{}), or \nDistRelMed{} of the run. Figure~\ref{fig:results}a compares the position of our origins with that of the corpus' decisive steps. In \nGoldSame{} of \nGoldN{} runs (\nGoldSamePct\%) the two coincide and in \nGoldWOne{} (\nGoldWOnePct\%) they are within one step; in \nGoldEarlier{} runs our origin is earlier, by a median of \nGoldGapMed{} steps, and in \nGoldLater{} it is later. Agreement within one step is higher on the short algorithm-generated runs (\nGoldAgWOnePct\%) than on the long hand-crafted ones (\nGoldHcWOnePct\%). When we are earlier, it is usually because the plan or the fact sheet already contained the wrong assumption and the decisive step merely acted on it. When we are later, the labelled step was in most cases a genuine error that the agents went on to repair, and a different fault, never corrected, is what reached the failure. We checked each of the \nGoldLater{} cases by hand.

\paragraph{Dynamics.} Contamination (H2: the deviated state is retained and reused) appears in 74 runs, observability debt (H6) in 44, repeated exposure (H1) in 38, drift (H3) in 36, cascade (H4) in 23 and budget exhaustion (H7) in 18. There is no external impact (H5) anywhere in the sample: in \nExtNone{} runs no action touched a shared or third-party system. This says something about the corpora, not about agents. A second, causally independent fault that also reached the failure was recorded in \nSecondary{} runs.

\subsection{RQ2: Evidence, detection, and response}
\label{sec:res-evidence}

In \nEvYes{} runs (\nEvYesPct\%) the trajectory contains, before the symptom, an observation or message that contradicts the erroneous belief or action; in \nEvBefore{} of them the evidence was already present when the fault entered. Agents acted on it in only \nEvUsed{} runs. In \nEvSeen{} the evidence was in an agent's input and was never mentioned; in \nEvAck{} an agent explicitly acknowledged it and continued unchanged, as when a verifier noted that distances were ``simulated'' and still reported the gyms as verified (Figure~\ref{fig:results}b). So in \nEvUnusedPct\% of the runs that had evidence, the run failed with the evidence needed to stop it sitting in the trajectory.

Some agent, tool, or verifier flagged a deviation in \nDet{} runs (\nDetPct\%): the erring agent or its orchestrator itself in 17 (for example a progress ledger noting a loop), a tool or verifier in 9, and another agent in 3. In a further 14 runs an agent only voiced uncertainty (``please verify'') and nobody acted on it. A response followed in \nResp{} runs, usually a replan or retry. Not one of these responses removed the deviated state; \nRespPartial{} removed part of it. The typical case is a replan that changes the next action and keeps the belief that caused the loop, so the loop comes back in a different form.

\subsection{RQ3: Which controls reach the episodes}
\label{sec:res-controls}

Table~\ref{tab:controlfreq} lists, for each control family of Table~\ref{tab:controlcontracts}, how often it was judged the most direct control and how often it would have reached the episode at all. An independent judge that reads the evidence the worker did not use reaches the most episodes (74\%), which is the evidence result of the previous section seen from the control side. Task contracts and scope rules fixed at admission (67\%), environment re-verification of beliefs before acting (43\%), stall and termination rules (40\%) and provenance on stored facts (38\%) follow. As the single most direct control, provenance comes first (20 runs); these are mostly the fabrications, which a rule of the form ``no unsourced claim is stored'' would have rejected.

Restoring the agent's own state to just before the origin would have sufficed in \nRestoreSuff{} runs. In \nRestorePlan{} runs the fault lay in the plan, the task brief, or a stop rule, so a restore would replay it; the remaining runs had contaminated shared state that a local restore does not reach.

\begin{table}[t]
\centering\small
\caption{Control families that would have observed and reached the episode. \emph{Primary}: judged the most direct control; \emph{Any}: listed among reaching controls (multi-label). Codes follow Table~\ref{tab:controlcontracts}: TC task contract, POL runtime policy, IFS information-flow separation, EVS environment-verified state, PROV provenance, TEST tests/oracles, JUDGE independent judge, CKPT checkpoint/restore, COORD coordination protocol, STALL stall/termination rule, HUMAN human review.}
\label{tab:controlfreq}
\begin{tabular}{@{}lrrr@{}}
\toprule Control&Primary&Any&Any share\\\midrule
TC&17&64&67\%\\
POL&3&30&32\%\\
IFS&1&4&4\%\\
EVS&17&41&43\%\\
PROV&20&36&38\%\\
TEST&8&20&21\%\\
JUDGE&11&70&74\%\\
CKPT&0&1&1\%\\
COORD&4&18&19\%\\
STALL&14&38&40\%\\
HUMAN&0&31&33\%\\
\bottomrule
\end{tabular}
\end{table}

\subsection{Propagation structure and differences between systems}
\label{sec:res-propagation}

An episode has a median of \nHopsMed{} recorded hops between origin and symptom (range \nHopsMin--\nHopsMax). The deviated state stays within one domain in \nDomOne{} runs, crosses two domains in \nDomTwo{}, and three or more in \nDomThreePlus{}; the common path is a belief formed in task state (R2) that is sent to another agent (R5) and then acted on (R3 or R4). On Who\&When, the median origin-to-symptom distance is \nDistDecision{} steps for decision faults ($n=\nCountDecision$), \nDistFabrication{} for fabrications ($n=\nCountFabrication$), and \nDistIntegration{} for misintegrated observations ($n=\nCountIntegration$): a fabricated fact is usually accepted and reported within a few turns, whereas a misread observation tends to be carried further before it surfaces. In \nSecIndep{} runs a second, causally independent fault also reached the failure, and in \nSecHarness{} runs the secondary event was a harness fault (Section~\ref{sec:res-harness}). Of the \nNfail{} runs, \nWrongOutput{} ended with a wrong output and \nNoOutput{} with no answer to the task (stall, truncation, or budget exhaustion).

Table~\ref{tab:corpus} splits the main measures by corpus. The two Who\&When systems fail in different ways. Magentic-One runs are long, almost always contain unused contradicting evidence (90\%), and in 55\% end with a harness fallback answer; their origin is rarely a fabrication (7\%). CaptainAgent runs are short, fabricate in 39\% of cases, and in 20\% fail already at admission because the manager handed the group a wrong task. Detection is more frequent in Magentic-One (41\%), because its orchestrator ledger flags loops, but as Section~\ref{sec:res-evidence} shows this detection did not produce effective responses. The MAD rows are too few to compare and are there for completeness.

\begin{table}[t]
\centering\scriptsize
\setlength{\tabcolsep}{3pt}
\caption{Main measures by corpus. Shares are percentages of the failed runs in each row: origin in P1--P2, origin kind fabrication, origin at admission (P0), contradicting evidence present, deviation detected, and harness fallback answer. Dist.\ is the median origin-to-symptom distance in steps (not comparable for MAD, whose traces are line-numbered logs).}
\label{tab:corpus}
\begin{tabular}{@{}lrrrrrrrr@{}}
\toprule Corpus (system)&$n$&P1--P2&Fabr.&P0&Evid.&Det.&Dist.&Harness\\\midrule
W\&W hand-crafted (Magentic-One)&29&52&7&0&90&41&18&55\\
W\&W algorithm-gen.\ (CaptainAgent)&59&66&39&20&64&25&5&0\\
MAD (mixed)&7&57&0&14&57&29&--&0\\
\bottomrule
\end{tabular}
\end{table}

\subsection{Symptom labels and origins}
\label{sec:res-mast}

For every run we also assigned the MAST failure modes~\citep{cemri2025mast} that describe the symptom (\nModesPerRun{} modes per run on average). Table~\ref{tab:mastcross} relates the frequent modes to the origin cells of the same runs. Each mode maps to several origin cells, between 4 and 11 distinct phase--domain pairs, and to several origin kinds. \emph{Incorrect verification} (3.3), for instance, covers \nMastIncVerKinds{} origin kinds: a verifier that rubber-stamps a fabricated fact, one that re-derives a wrong decision from the same inputs, and one that validates a solution to the wrong task after a corrupted handoff. The mapping is also many-to-one in the other direction: the P1/R2 cell (a wrong deliberation over task state) appears among the three most frequent cells of every one of the ten modes. So a mode label does not tell you at which phase a control would have to act, nor which state it would have to inspect. That is the part the episode record supplies.

\begin{table}[t]
\centering\scriptsize
\setlength{\tabcolsep}{3pt}
\caption{MAST symptom modes assigned to the same runs, with the number of distinct origin cells (phase/domain) and the most frequent cells and origin kinds. Modes with fewer than five runs are omitted.}
\label{tab:mastcross}
\begin{tabularx}{\linewidth}{@{}P{.3\linewidth}rrXP{.24\linewidth}@{}}
\toprule MAST mode&Runs&Cells&Most frequent origin cells&Origin kinds\\\midrule
1.1 Disobey task specification&49&11&P1/R2 (14), P2/R2 (12), P0/R5 (6)&decision 23, fabrication 15, coordination 5\\
1.2 Disobey role specification&7&4&P2/R2 (3), P1/R2 (2), P2/R3 (1)&fabrication 3, decision 3, interface 1\\
1.3 Step repetition&33&7&P1/R2 (10), P2/R2 (8), P3/R2 (6)&decision 17, fabrication 7, integration 6\\
1.5 Unaware of stopping conditions&5&4&P1/R2 (2), P2/R3 (1), P2/R2 (1)&decision 3, interface 1, coordination 1\\
2.2 Fail to ask for clarification&12&5&P0/R5 (6), P1/R2 (3), P2/R2 (1)&coordination 5, decision 4, specification 2\\
2.3 Task derailment&15&6&P0/R5 (9), P1/R2 (2), P0/R3 (1)&coordination 6, specification 3, decision 3\\
2.6 Reasoning--action mismatch&43&10&P2/R2 (14), P1/R2 (11), P2/R3 (5)&decision 18, fabrication 15, integration 4\\
3.1 Premature termination&22&11&P1/R2 (6), P4/R2 (3), P2/R2 (2)&decision 8, interface 4, fabrication 3\\
3.2 No or incomplete verification&44&10&P1/R2 (14), P2/R2 (10), P3/R2 (8)&decision 15, fabrication 14, integration 8\\
3.3 Incorrect verification&44&8&P2/R2 (17), P0/R5 (8), P1/R2 (6)&decision 16, fabrication 16, coordination 6\\
\bottomrule
\end{tabularx}
\end{table}

\subsection{A harness-level failure hidden in attribution labels}
\label{sec:res-harness}

This one we did not go looking for. While annotating the Magentic-One runs we kept seeing the same ending: an HTTP~400 from the model provider's content filter on the orchestrator's progress-ledger call, after which the experiment script printed a \texttt{FINAL ANSWER} that no agent had produced. A deterministic scan of all raw Who\&When trajectories finds this crash in \nWWhcCrash{} of \nWWhcN{} hand-crafted runs (\nWWhcCrashPct\%), \nWWhcFallback{} of which end with such a fallback answer, and in \nWWagCrash{} of \nWWagN{} algorithm-generated runs. For the \nWWhcFallback{} fallback runs the corpus attributes the failure to the orchestrator in \nFallbackBlameOrch{} cases and to another agent (mostly the web surfer) in \nFallbackBlameOther{}. In our records the event is a P6 termination fault in the interface domain (R3), the kind of fault that neither an agent-level symptom taxonomy nor a responsible-agent label represents. Attribution methods evaluated on this split are therefore partly scored on runs whose final answer came from the harness, not from any agent.

\section{Discussion}
\label{sec:discussion}

\paragraph{What the record adds to a label.} The clearest illustration is MAST mode 3.3, \emph{incorrect verification}. We assigned it to 44 runs, and those runs have five different kinds of origin: 16 wrong decisions, 16 fabrications, 6 corrupted handoffs, 4 misread observations, 2 defective briefs. The verifier rubber-stamped something in all of them, but what it should have checked differs completely: the provenance of a claimed fact, the environment state behind a belief, or whether the task it was verifying was the task the user asked. A label says ``the verifier failed''; the record says what it would have had to look at. The comparison with the Who\&When decisive step says the same thing from the other side. The step a human picks as decisive is often downstream of where the wrong belief was first written into a plan or a fact sheet.

\paragraph{Evidence is there; nobody reads it.} If one number from this study should be remembered, it is that \nEvYesPct\% of failed runs contain, before the failure, something in the trajectory that contradicts the error, and that in \nEvUnusedPct\% of those cases no agent did anything with it. This is consistent with reports that agents rarely catch their own mistakes~\citep{rahman2026traverse} and that a verifier sharing the worker's information adds little~\citep{huang2024selfcorrect,zanolini2026correlatedmonitors}. What our data add is a target: a control earns its place if it reads the evidence the worker left on the table. Independent judges and environment re-verification do this by construction, and they are the two families that reach most of our episodes.

\paragraph{Detection is not the problem either.} Magentic-One's orchestrator has loop detection, and it fired in a good share of the runs. The replans that followed changed the next action and kept the belief that had caused the loop, so the loop came back in a different form. Recovery that does not revise the contaminated state reproduces the failure. This is also why a local checkpoint would have helped in a minority of runs only: in most, replaying from the checkpoint would replay the plan.

\paragraph{Harness faults belong in the denominator.} The content-filter crashes were a surprise. Almost half of one benchmark split ends with the harness, not an agent, producing the answer that gets scored. Attribution benchmarks should mark such runs, and evaluations of agents should report harness-terminated runs separately from agent errors; otherwise a method that learns to blame the web surfer for a provider's HTTP 400 will look good.

\subsection{Threats to validity}
\label{sec:limitations-study}

\paragraph{Annotator.} The records were produced by an LLM following a written protocol. We have no human inter-annotator agreement to report in this version. What we have instead: every step field is anchored to a line in the trace and can be checked; records passed mechanical schema validation; our origins agree with the human decisive steps within one step in \nGoldWOnePct\% of cases and disagree in the direction of ``later'' in only \nGoldLater{} of \nGoldN{}; and the harness-fault result is independent of annotation. A human check of a subset is the obvious next step, and we would not be surprised if it moved some of the kind labels, which were the hardest part of the protocol.

\paragraph{Sample.} \nNann{} annotated runs out of 203 normalized traces, selected by annotation batch rather than at random. The Who\&When part matches the corpus in run length. The MAD part is small and short-biased, and we only show it for completeness.

\paragraph{Corpora.} Both corpora are question answering and programming in sandboxes. Not one run in our sample changes a shared or third-party system, so external impact (H5) and recovery of committed effects, which the framework was partly built for, cannot be studied here. That needs corpora like $\tau$-bench or TheAgentCompany trajectories, or fault injection.

\paragraph{Counterfactual controls.} ``Would have reached'' is a judgment about the trace, constrained by the control contracts of Table~\ref{tab:controlcontracts}. It is not a measured intervention. Whether a control would actually have prevented a given failure requires re-running the agents with the control in place, which we have not done.

\section{Conclusion}
\label{sec:conclusion}

We took an episode-level record of faults, wrote it down as an annotation protocol, and applied it to \nNfail{} released failures of multi-agent LLM systems. Faults enter early, mostly as decisions or fabrications over task state, and often before the step that human labels call decisive. In most failed runs the trajectory already contained the evidence that would have stopped the episode, and nobody used it. Detection happened in a third of the runs and never led to a response that removed the deviated state. Nearly half of one benchmark split was terminated by the harness rather than by any agent. The controls that reach these episodes are the ones that read evidence the worker ignored, fix the task contract at admission, and re-check beliefs against the environment; local checkpointing reaches few of them. The protocol, the records and the scripts are included so the analysis can be repeated, extended to new corpora, and checked against human annotation.

\begin{acks}
AI-assisted tools were used for the annotation described in Section~\ref{sec:procedure}, and for editing, source checking and figure preparation. The authors are responsible for all claims and interpretations.
\end{acks}

\bibliographystyle{ACM-Reference-Format-unsrt}
\bibliography{references}

\appendix
\section{Background: Dynamics, Controls, and Published Evidence}
\label{sec:background}
The appendices below give the background the episode framework rests on: the long-horizon dynamics it encodes (Appendix~\ref{sec:challenges}), the control families and their contracts used in Table~\ref{tab:controlfreq} (Appendix~\ref{sec:mechanisms}), evaluation practice (Appendix~\ref{sec:evaluation}), and a synthesis of published evidence (Appendix~\ref{sec:open}).
\section{Challenges and Long-Horizon Dynamics}
\label{sec:challenges}

Because each decision in agent execution carries state and consequences forward, six challenges arise. Sections~\ref{sec:ch1}--\ref{sec:ch2} examine dependent errors and persistent state, Sections~\ref{sec:ch3}--\ref{sec:ch4} actions and incomplete verification, and Sections~\ref{sec:ch5}--\ref{sec:ch6} coordination and oversight, while Section~\ref{sec:ch7} consolidates them into seven dynamics.

\subsection{Repeated execution and dependent errors}
\label{sec:ch1}

This subsection asks how the chance of failure changes as a run gets longer and after an error has occurred. A familiar illustration assumes $N$ necessary steps, each independently correct with probability $p$:
\begin{equation}
\Pr(\text{all steps correct})=p^N.
\label{eq:bernoulli}
\end{equation}
At $p=.99$, this probability falls below one half at 69 steps. This is only a baseline, because recovery allows success after an error and dependence can push joint correctness well above or below $p^N$. For an absorbing service-failure event $F$, the chain rule gives the general form
\begin{equation}
S_N=\Pr(F\text{ absent by }N)=\prod_{t=1}^{N}(1-\lambda_t),\qquad
\lambda_t=\Pr(F\text{ first occurs at }t\mid F\text{ absent before }t),
\label{eq:hazard}
\end{equation}
where $\lambda_t$ is the marginal hazard, averaged over histories surviving to $t$. The path-level hazard $\lambda_t(H_{t-1})$ may depend on the task, preceding history $H_{t-1}$, and active controls, so the empirical question is how history shapes it.

Several results are consistent with hazards that rise after errors. Closest to an intervention, Sinha et al. supply knowledge and plans, then show that injecting erroneous earlier outputs into history reduces later accuracy~\citep{sinha2025illusion}. In repeated Toolathlon runs, controlling for model and task, an off-canonical call is associated with a 22.7-point increase in next-call deviation~\citep{lee2026canonical}. SWE-agent's GPT-4 edits eventually succeed 90.5\% of the time, but only 57.2\% after a failed edit~\citep{yang2024sweagent}. At larger scale, every Vending-Bench model had runs that derailed, and breakdowns did not coincide with context-window filling~\citep{backlund2025vendingbench}. Across 1,794 coding-agent trajectories, annotated failures were predominantly epistemic, with early onset and late-surfacing symptoms~\citep{zhao2026failureprocess}. Across 2,518 TRAVERSE trajectories, agents often failed to recover from a first mistake and rarely caught their own errors~\citep{rahman2026traverse}. The injected-history design and a natural experiment holding model and task fixed~\citep{lee2026canonical} support causal claims. The rest are observational, and because errors are not randomly assigned, task difficulty may partly explain their associations; in pooled failure-prediction data, almost every pair behind the reported AUROC compares runs of different tasks, and within-task discrimination early in a run is near chance~\citep{esfandyari2026taskdifficulty}.

\begin{figure}[tbp]
\centering\includegraphics[width=.98\linewidth]{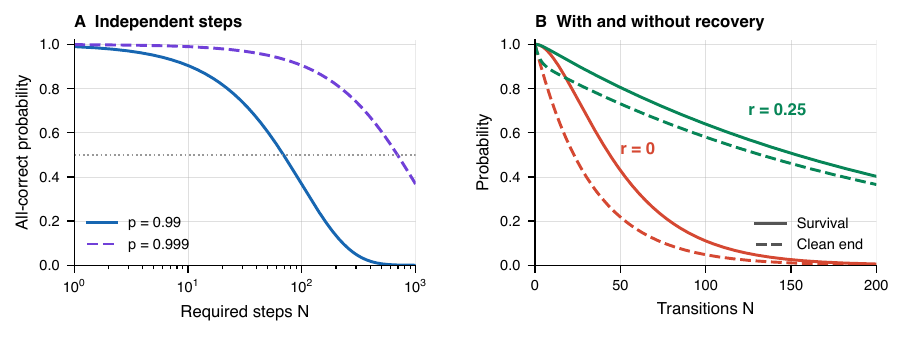}
\caption{Computed illustrations, not empirical results. Left: all-correct probability under Equation~\eqref{eq:bernoulli}. Right: survival and clean-endpoint probability under Equation~\eqref{eq:recoverymodel} with $q=.03$, $b=.05$, and recovery probability $r$ labeled on the curves.}
\label{fig:reliability}
\Description{Two analytical plots: independent step correctness decreases with required steps; in a three-state model, increasing recovery improves survival, while clean endpoints are less frequent than survival.}
\end{figure}

A three-state model with illustrative parameters shows why detection and recovery matter. Let states $\mathsf{c},\mathsf{x},\mathsf{f}$ denote clean, contaminated, and irreversibly failed execution, with unit vectors $e_{\mathsf{c}},e_{\mathsf{x}}$:
\begin{equation}
K=\begin{pmatrix}1-q&q&0\\r&1-r-b&b\\0&0&1\end{pmatrix},\quad q\in[0,1],\ r,b\geq0,\ r+b\leq1,
\label{eq:recoverymodel}
\end{equation}
where $q$ introduces an error, $r$ recovers, and $b$ turns contamination into failure. Early, hidden errors correspond to a small detection-driven $r$, and meltdown loops to a rarely left contaminated state. With $q=.03$ and $b=.05$, raising $r$ from 0 to $.25$ raises survival $e_{\mathsf{c}}^\top K^N(e_{\mathsf{c}}+e_{\mathsf{x}})$ at $N=100$ from about 0.1 to about 0.6, and the clean endpoint $e_{\mathsf{c}}^\top K^Ne_{\mathsf{c}}$ from about 0.05 to about 0.6 (Figure~\ref{fig:reliability}). Long-run survival in this model thus depends mainly on the detection-driven recovery rate $r$.

\subsection{Persistent state and contaminated evidence}
\label{sec:ch2}

This subsection turns to the state an agent carries between steps: context, memory, and stored evidence. Agents can fail to \emph{use} evidence, not only \emph{lose} it through truncation, summarization, or cache compression. For example, relevant information placed in the middle of a long prompt can go unused~\citep{liu2024lostmiddle}, while retrieval and external memory~\citep{lewis2020rag,packer2023memgpt,shinn2023reflexion} add further problems of selection, freshness, and retraction. Planning benchmarks show what losing constraints costs. In TravelPlanner, GPT-4 reached a 0.6\% final success rate, failing to stay on task, choose tools, and track multiple constraints~\citep{xie2024travelplanner}. MemoryAgentBench separately evaluates retrieval, test-time learning, long-range understanding, and selective forgetting, and no evaluated memory design masters all four~\citep{hu2026memoryagentbench}. Stored evidence should therefore keep its source, version, and uncertainty through summarization, because otherwise a summary that turns ``not checked'' into ``verified'' silently changes the item's meaning (Section~\ref{sec:mech2}).

\subsection{Actions, trust boundaries, and external impact}
\label{sec:ch3}

This subsection concerns actions, the point where an agent's decision becomes an effect in a system it does not control. A valid tool call can be the wrong action, so tool-learning benchmarks separate tool-selection errors from invocation errors~\citep{qin2024toolbench}. Emulation and injection studies show harmful actions can also be syntactically ordinary~\citep{ruan2024toolemu,debenedetti2024agentdojo}. Indirect prompt injection crosses a central boundary between \emph{data} that informs a decision and \emph{authority} that permits an effect, along three paths: instructions embedded in retrieved content~\citep{greshake2023greshake}, poisoned memory that persists across sessions, as in AgentPoison~\citep{chen2024agentpoison}, and tool ecosystems, where Model Context Protocol (MCP) servers present 16 lifecycle threat scenarios~\citep{hou2025mcp} and MCP tools can coerce leading models into malicious code execution and credential theft~\citep{radosevich2025mcpaudit}. Measured attack success depends on the threat model. InjecAgent reports 24\% for its ReAct-prompted GPT-4 setting over 1,054 cases~\citep{shi2024injecagent}, whereas AgentDojo pairs 97 legitimate tasks with 629 security cases~\citep{debenedetti2024agentdojo}.

Recovery after an action takes effect depends on deployment: a test database can be reset, but a sent message stays sent. Silence is not evidence of failure, because a lost response may follow a successful commit. Retry safety therefore requires execution evidence, deduplication, or reconciliation~\citep{wu2026saferesume,zheng2026acrfence}. Two 2026 benchmarks measure the cost of getting this wrong: in LIMBO, frontier models duplicate an in-flight or redelivered write in 56--74\% of episodes, and an idempotency key on every write lowers the duplicate rate from 28\% to 4\%~\citep{li2026limbo}; in UndoBench, naive retry after a lost acknowledgment duplicated external effects in 53\% of trials~\citep{sah2026undobench}.

\subsection{Verification with incomplete or dependent evidence}
\label{sec:ch4}

This subsection considers what a check can establish when it shares evidence with the work it checks. Verification can target action validity, state fidelity, progress, or completion; a correct final state can coexist with harmful intermediate actions. The central difficulty is that verifiers often share the worker's information. Intrinsic self-correction can fail to improve or even degrade reasoning~\citep{huang2024selfcorrect}, whereas Self-Refine and chain-of-verification gain when feedback or independent questions are available~\citep{madaan2023selfrefine,dhuliawala2024chainverif}. Diagnosing agent traces is itself hard: on the 148 annotated traces of TRAIL (Trace Reasoning and Agentic Issue Localization), the best long-context model scored 11\%~\citep{deshpande2025trail}. Agents also misreport their own checking: coding agents skipped files they were asked to review in 68\% of runs and, in 80\% of those, claimed a complete review or left the gap undisclosed~\citep{smyth2026overclaim}; under injected tool faults, agents flagged an explicit error in 91\% of trials but a plausible wrong value in only 59\%~\citep{kraishan2026loudquiet}. Given shared evidence, verification claims should state what was checked and where, separately from confidence and correctness.

\subsection{Coordination and common-cause failure}
\label{sec:ch5}

This subsection looks at errors that arise between agents rather than within one. Multi-agent frameworks distribute work through roles, messages, and shared artifacts~\citep{hong2024metagpt,qian2024chatdev,wu2023autogen,li2023camel,chen2024agentverse}, which improves decomposition but introduces stale reads, conflicting writes, and unclear ownership. An enforced access rule prevents one agent from overwriting another's work, although it protects only the operations it covers. Fault injection into inter-agent messages shows how far errors travel: across 29,000 executions, 21\% of failures produced no immediate signal, and a flat topology kept 53\% task success under message faults where sequential and hierarchical topologies kept about 3\%, because an alternative path preserved the evidence~\citep{zhou2026messagesgowrong}. Adversarial content in shared artifacts reached 60--80\% of agents over up to eight handoffs~\citep{pulipaka2026shareborne}, and CaMeL's per-agent guarantee does not compose when one agent calls another as a tool~\citep{petersgill2026camelstalk}. Classical results also limit coordination: FLP rules out guaranteed deterministic consensus in an asynchronous system with even one crash fault~\citep{fischer1985flp}, so practical protocols assume partial synchrony~\citep{castro1999pbft}.

\subsection{Human oversight and limited attention}
\label{sec:ch6}

The last challenge is the human in the loop. Oversight adds judgment and authority, but its capacity depends on reviewer workload and available evidence. Pan et al.'s study of 86 deployed systems across 26 domains (MAP) found that 68\% of systems run at most ten steps before intervention and 74\% rely primarily on human evaluation~\citep{pan2025map}. READY makes oversight cost measurable: two systems 0.3 points apart in autonomous accuracy required 39.2\% versus 29.6\% human review to reach the same 76\% reliability target~\citep{chatrath2026ready}. Programming-assistant studies show the same need to measure review effort~\citep{vaithilingam2022expectation}, and in a controlled study of 36 participants overseeing a coding agent, people were confident while missing silent defects, because they ran generic checks rather than generating diagnostic evidence~\citep{dissanayake2026looksfine}.

\subsection{Seven dynamics}
\label{sec:ch7}

Table~\ref{tab:dynamics} consolidates the challenges into seven overlapping dynamics, and the evidence above supports \emph{trajectory amplification} as a working hypothesis: error rates rise after earlier errors, onset is early while symptoms appear late, and derailment is unrelated to context capacity.

\begin{table}[t]
\centering\small
\caption{Seven analytical dynamics.}
\label{tab:dynamics}
\begin{tabularx}{\linewidth}{@{}P{.05\linewidth}P{.22\linewidth}P{.31\linewidth}X@{}}
\toprule\textbf{Code}&\textbf{Dynamic}&\textbf{Distinguishing observation}&\textbf{Useful measurement}\\\midrule
H1&Repeated exposure&More opportunities for an event.&Hazard per opportunity; task failure.\\
H2&Contamination&A deviated state is retained and reused.&Retention, consumption, retraction.\\
H3&Drift&Divergence from an explicit reference.&Goal, plan, or policy deviation.\\
H4&Cascade&One error influences several later events.&Supported dependency edges.\\
H5&External impact&Restoration omits consequences.&Unrecovered effects and cost.\\
H6&Observability debt&Diagnosis needs missing dependencies.&Detection delay, attribution error.\\
H7&Resource exhaustion&Execution or recovery exceeds budget.&First exhausted resource.\\
\bottomrule
\end{tabularx}

\end{table}

To test whether the framework can express an independently derived taxonomy, Table~\ref{tab:mast} maps the 14 failure modes of MAST~\citep{cemri2025mast} onto it. Every mode maps onto at least one phase--domain cell, and modes that MAST lists separately can share a coordinate. For example, task derailment and reasoning--action mismatch are both P1/R2 deviations from an explicit reference (H3). MAST's conversational focus excludes model and compute state (R1), external impact (H5), and recovery of committed effects. Eight of the 14 modes have no intrinsic dynamic. For instance, premature termination, a termination-decision error (P4--P6), implies exhaustion when a budget forced the stop, while incorrect verification causes contamination when the accepted output is reused. Section~\ref{sec:res-mast} tests this mapping empirically. The mapping is our interpretation of MAST's published definitions.

\begin{table}[!tbp]
\centering\footnotesize
\setlength{\tabcolsep}{4pt}
\caption{Crosswalk from MAST's failure taxonomy to the framework. No mode falls in R1 or H5, none concerns recovery of external effects, and eight modes have no intrinsic dynamic.}
\label{tab:mast}
\begin{tabularx}{\linewidth}{@{}P{.18\linewidth}X@{}}
\toprule\textbf{MAST category}&\textbf{Failure modes with framework coordinates (phase / domain / dynamic; parenthesized dynamics apply only under the stated condition)}\\\midrule
Specification / system design&1.1 Disobey task specification (P0--P1 / R2 / H3); 1.2 Disobey role specification (P1 / R5 / H3); 1.3 Step repetition (P1--P2 / R2 / H1; H7 if a budget is exhausted); 1.4 Loss of conversation history (P3 / R2 / --; H3 if later divergence follows); 1.5 Unaware of stopping conditions (P4--P6 / R2 / --; H7 if continuation exhausts a budget)\\
Inter-agent misalignment&2.1 Conversation reset (P5 / R5 / --; state loss); 2.2 Fail to ask for clarification (P1, P5 / R5--R6 / --; H3 if the unresolved ambiguity causes divergence); 2.3 Task derailment (P1 / R2 / H3); 2.4 Information withholding (P6 / R5 / H6); 2.5 Ignored other agent's input (P3 / R5 / --; H4 if downstream influence is shown); 2.6 Reasoning--action mismatch (P1--P2 / R2 / H3)\\
Task verification&3.1 Premature termination (P4--P6 / R2 / --; H7 only if a budget forced termination); 3.2 No or incomplete verification (P4 / R2, R4 / --; H6 if later diagnosis needs the unchecked evidence); 3.3 Incorrect verification (P4 / R2 / --; H2 if the accepted output is reused)\\
\bottomrule
\end{tabularx}

\end{table}

\section{Mechanisms: What They Control and What They Assume}
\label{sec:mechanisms}

Established reliability methods work precisely where their assumptions hold for the agent, so we compare controls by what each establishes and assumes. Planning and memory support consistent decisions (Sections~\ref{sec:mech1}--\ref{sec:mech2}), while enforcement, recovery, and verification limit incorrect effects (Sections~\ref{sec:mech3}--\ref{sec:mech4}), and coordination and oversight govern shared work (Sections~\ref{sec:mech5}--\ref{sec:mech6}). Section~\ref{sec:mech7} compares all families and their composition.

\subsection{Planning, decomposition, and learned execution}
\label{sec:mech1}

\paragraph{Deliberation and search.} These methods address how agents reason and plan before acting. Chain-of-thought and self-consistency help where intermediate reasoning or sampled answers are informative~\citep{wei2022cot,wang2023selfconsistency}, and ReAct couples deliberation to observations~\citep{yao2023react}. ReWOO separates planning from execution and reduces model calls~\citep{xu2024rewoo}. Tree of Thoughts and LATS search with evaluation and backtracking~\citep{yao2023tot,zhou2023lats}, while SayCan ties plans to feasible actions~\citep{ahn2022saycan}. LLM-Modulo delegates plan validity to external verifiers~\citep{kambhampati2024llmmodulo}. RAP plans with an LLM world model~\citep{hao2023rap}. On long tasks, plans go stale, replanning follows faulty observations, search depends on its evaluator, and planning on its model.

\paragraph{Revision, skills, and harness structure.} These methods add revision, retained experience or control structure. Self-Refine revises answers, Reflexion retains episodic feedback, and Voyager accumulates executable skills~\citep{madaan2023selfrefine,shinn2023reflexion,wang2024voyager}. A skill with a testable postcondition is a stronger contract than an unverified lesson, because it can be checked. Revision needs discriminating feedback, or iteration can reinforce a wrong diagnosis~\citep{huang2024selfcorrect}. CRITIC grounds revision in tool feedback, and process supervision evaluates intermediate steps~\citep{gou2024critic,lightman2024letsverify}. Decomposition with local error detection and voting sustains a million-step task~\citep{meyerson2025million}, while fixed workflows such as Agentless show that simpler control can be preferable~\citep{xia2025agentless}. LongHorizon-Harness stores task state outside the model's context, updates it with environment-verified facts, and reports gains on Terminal-Bench and OSWorld~\citep{ma2026longhorizonharness}. This trades a contamination risk in task state (H2, R2) for an environment check (P4, R4) performed before each planning decision.

\paragraph{Training and admission.} These methods train agents for long tasks. WebRL trains web agents with a curriculum built from failed attempts and an outcome reward model~\citep{qi2025webrl}. SWE-RL uses a rule-based similarity reward on software-evolution data~\citep{wei2025swerl}, and AREX combines constraint-wise verification with long-horizon training~\citep{lu2026arex}. Such rewards can improve completion yet leave unrewarded properties, such as avoiding harmful intermediate actions, to runtime checks. A task contract fixed at admission (P0) records outcome, non-goals, evidence, limits and authority, so acceptance criteria stay unchanged as the plan evolves.

\subsection{Memory and compute-state fidelity}
\label{sec:mech2}

\begin{figure}[tbp]
\centering\resizebox{.98\linewidth}{!}{\definecolor{smA}{HTML}{7040D8}\definecolor{smB}{HTML}{1766B1}\definecolor{smC}{HTML}{078557}\definecolor{smD}{HTML}{D54832}\definecolor{smInk}{HTML}{20242A}\definecolor{smG}{HTML}{6B6F76}
\begin{tikzpicture}[x=1cm,y=1cm,font=\sffamily\small,
 hd/.style={rounded corners=3pt,fill=#1,text=white,font=\sffamily\normalsize\bfseries,text width=3.1cm,minimum height=.95cm,align=center},
 bx/.style={draw=#1,fill=#1!5,rounded corners=3pt,line width=.6pt,align=center,text width=3.5cm,minimum height=.95cm,text=#1!75!black},
 hist/.style={align=center,text width=2.5cm,text=smInk},
 chk/.style={align=center,text width=2.7cm,text=smInk},
 ar/.style={-{Stealth[length=4pt]},line width=.6pt,draw=smInk!70}]
\foreach \y/\c/\h/\s/\p/\q/\k/\rc in {
 0/smB/Exact KV storage/{History:\\A\,B\,C\,D}/{\textbf{Move physical pages}\\C\,D\,$|$\,A\,B}/{\textbf{Keep logical access}\\A\,B\,C\,D}/{\textbf{Check:}\\values and identity}/smC,
 -1.3/smA/Selective retention/{History:\\A\,\textcolor{smD}{\textbf{B}}\,C\,D}/{\textbf{Keep selected history}\\A\,C\,D}/{\textbf{B is missing}\\a later action needs B}/{\textbf{Check:}\\later decisions}/smD,
 -2.6/smC/Semantic memory/{Source:\\an uncertain claim}/{\textbf{Store or condense}\\a claim, lesson, or skill}/{\textbf{Meaning may change}\\source or uncertainty lost}/{\textbf{Check:}\\source and updates}/smD}{
 \node[hd=\c] (h) at (1.65,\y) {\h};
 \node[hist] (s) at (4.65,\y) {\s};
 \node[bx=\c] (a) at (7.95,\y) {\p};
 \node[bx=\rc] (b) at (12.05,\y) {\q};
 \node[chk] (k) at (15.35,\y) {\k};
 \draw[ar] (s) -- (a); \draw[ar] (a) -- (b); \draw[ar,dashed] (b) -- (k);}
\end{tikzpicture}}
\caption{Memory operations differ in what they preserve: exact paging retains logical access, selective retention can remove a later-needed constraint, and semantic memory must preserve source applicability.}
\label{fig:memory}
\Description{Three panels contrast exact KV page management, selective loss of a required constraint, and external memory that may omit provenance or uncertainty.}
\end{figure}
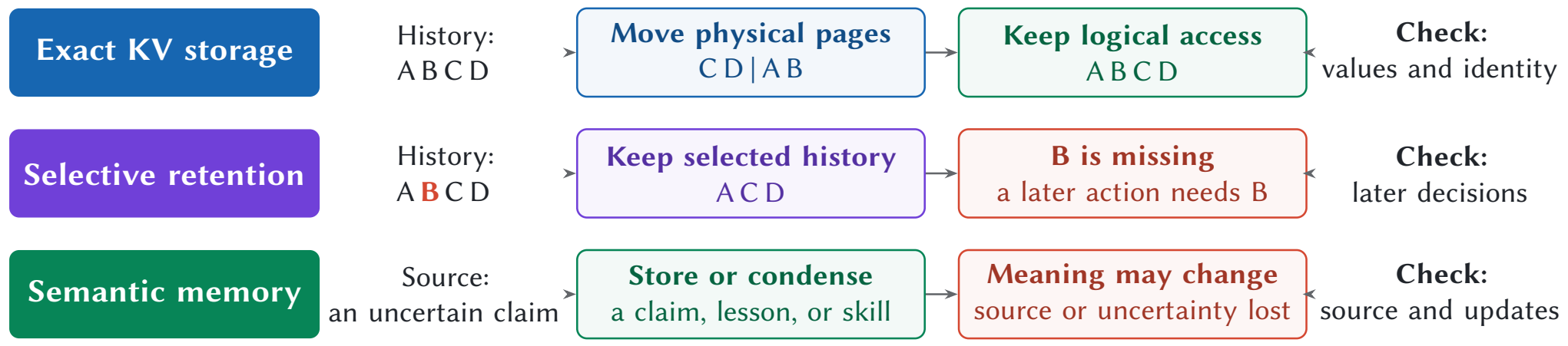

This subsection covers controls over what an agent remembers, from the KV cache to external memory. Memory operations differ in what they preserve, so each kind needs its own evaluation (Figure~\ref{fig:memory}).

\textbf{Exact operations.} PagedAttention changes where key--value (KV) cache entries are stored but preserves logical access, and FlashAttention computes exact attention up to floating-point reordering~\citep{kwon2023vllm,dao2022flashattention}. Divergence beyond floating-point nondeterminism therefore indicates an implementation or numerical fault, not lossy retention. \textbf{Lossy retention.} H$_2$O, Scissorhands, and SnapKV select which positions to retain~\citep{zhang2023h2o,liu2023scissorhands,li2024snapkv}, whereas KIVI quantizes stored values~\citep{liu2024kivi}. Their selection signals do not measure a task constraint's importance, so both approaches should be evaluated by their effect on later decisions under matched budgets. \textbf{Agent-context compression.} ACON refines natural-language compression guidelines from analyses of agent failures. This reduces peak tokens by 26--54\% while improving success on AppWorld, OfficeBench, and multi-objective QA~\citep{kang2026acon}. The result supports judging compression by its effect on later decisions.

External-memory systems each control a different aspect of memory. Retrieval-augmented generation (RAG) supplies non-parametric evidence~\citep{lewis2020rag}, while MemGPT manages memory tiers~\citep{packer2023memgpt}. Self-RAG and CRAG control retrieval timing and evidence quality~\citep{asai2024selfrag,yan2024crag}. Generative Agents, A-MEM, and Mem0 consolidate and link memories~\citep{park2023generativeagents,xu2025amem,chhikara2025memzero}. Stored facts need verification, so we propose two requirements. First, memory should record source versions, observation times, uncertainty, and dependent conclusions, so a changed fact invalidates the conclusions built on it. Second, instructions quoted in stored text should never trigger actions, because quoting an instruction does not authorize it.

For R1, hardware protection and fail-stop assumptions remain relevant~\citep{koren2007faulttol,schlichting1983failstop}. Fault-injection tools such as Ares and PyTorchFI measure how perturbations affect inference~\citep{reagen2018ares,mahmoud2020pytorchfi}. One study in our corpus follows such perturbations into agent loops: Rubey et al. inject random two-bit weight faults in 4,300 agent runs and find that agents often recover and complete the task, at higher cost, while a few faults produce semantically meaningful errors with otherwise normal behavior~\citep{rubey2026hardwarefaults}.

\subsection{Interface enforcement and recovery boundaries}
\label{sec:mech3}

This subsection covers controls at the interface between the agent and its environment, before and after an action takes effect. Valid format, permission, and reversibility are independent guarantees. Enforcement constrains an action before it occurs, whereas isolation, rollback, and compensation address its effects afterward.

\paragraph{Pre-execution schemas, policies, and guardrails.} Constrained decoding enforces grammars~\citep{willard2023grammar,beurerkellner2023lmql}, but a valid parse does not prove authorization. AgentSpec evaluates user rules at runtime triggers~\citep{agentspec2025}. CaMeL separates trusted control flow from untrusted data and checks capabilities~\citep{debenedetti2025camel}. Progent checks tool calls against symbolic privilege rules, narrowing automatically but requiring approval to expand~\citep{shi2025progent}. Llama Guard constrains conversational inputs and outputs~\citep{inan2023llamaguard}, whereas GuardAgent and ShieldAgent check proposed actions with generated code or policy-derived rule circuits~\citep{xiang2025guardagent,chen2025shieldagent}. Instruction Hierarchy trains models to resist injected instructions~\citep{wallace2024hierarchy}. Deterministic decisions are only as reliable as their rules, learned decisions need their error measured on the deployment distribution, and a check goes stale unless it is bound to the resource version it examined.

\paragraph{Post-execution isolation, restoration, and compensation.} MicroVMs limit reachable resources~\citep{agache2020firecracker}, emulated tools avoid real services~\citep{ruan2024toolemu}, and checkpoint/restart and message logging restore computational state~\citep{elnozahy2002rollback}. The question is what a snapshot contains and which effects escaped it. Retry, replan, rollback, compensation, and escalation differ in preconditions and residual risks. Retry is safe only with evidence of non-execution, idempotence, or deduplication. Compensation issues an offsetting action, as in sagas~\citep{garcia2004saga}, and can itself fail. Self-healing designs~\citep{ghosh2007selfhealing} also need versions, because a rewound agent can conflict with peers.

\begin{figure}[tbp]
\centering\resizebox{.98\linewidth}{!}{\definecolor{smA}{HTML}{7040D8}\definecolor{smB}{HTML}{1766B1}\definecolor{smC}{HTML}{078557}\definecolor{smD}{HTML}{D54832}\definecolor{smInk}{HTML}{20242A}\definecolor{smG}{HTML}{6B6F76}
\begin{tikzpicture}[x=1cm,y=1cm,font=\sffamily\small,
 hd/.style={rounded corners=3pt,fill=#1,text=white,font=\sffamily\normalsize\bfseries,text width=5.1cm,minimum height=.95cm,align=center},
 bx/.style={draw=#1,fill=#1!5,rounded corners=3pt,line width=.6pt,align=center,text width=3.7cm,minimum height=.95cm,text=#1!75!black},
 cv/.style={align=center,text width=2.5cm,text=smInk},
 ar/.style={-{Stealth[length=4pt]},line width=.6pt,draw=smInk!70},
 band/.style={rounded corners=3pt,fill=#1,text=white,font=\sffamily\normalsize\bfseries,minimum width=16.2cm,minimum height=.75cm,align=center}]
\node[band=smA] at (8.1,0) {One booking request, an unknown response};
\node[bx=smB] (s1) at (2.9,-1.2) {\textbf{Send request A17}\\agent calls the service};
\node[bx=smD] (s2) at (8.1,-1.2) {\textbf{Booking commits}\\provider records the effect};
\node[bx=smB] (s3) at (13.3,-1.2) {\textbf{Response times out}\\agent has no confirmation};
\draw[ar] (s1) -- (s2); \draw[ar] (s2) -- (s3);
\foreach \y/\c/\h/\p/\q/\v in {
 -2.85/smD/{A\quad Assume nothing happened}/{\textbf{Retry as a new request A18}\\the first booking\\may already exist}/{\textbf{Duplicate booking}\\the one-booking rule is broken}/{a client reset\\cannot undo\\a booking},
 -4.6/smC/{B\quad Check what happened}/{\textbf{Query status or deduplicate}\\use durable\\service evidence}/{\textbf{Reconcile or escalate}\\keep unresolved effects explicit}/{reconciliation\\needs service\\support}}{
 \node[hd=\c] (h) at (2.7,\y) {\h};
 \node[bx=\c] (a) at (7.45,\y) {\p};
 \node[bx=\c] (b) at (11.55,\y) {\q};
 \node[cv] (v) at (15.05,\y) {\v};
 \draw[ar] (h) -- (a); \draw[ar] (a) -- (b); \draw[ar,dashed] (b) -- (v);}
\end{tikzpicture}}
\caption{Two recovery paths after an uncertain response whose provider commitment the agent has not observed: blind retry creates a duplicate; reconciliation preserves uncertainty until service evidence is available.}
\label{fig:recoverypaths}
\Description{A committed booking whose response times out branches into blind retry and a duplicate, or status reconciliation and evidence-based recovery.}
\end{figure}
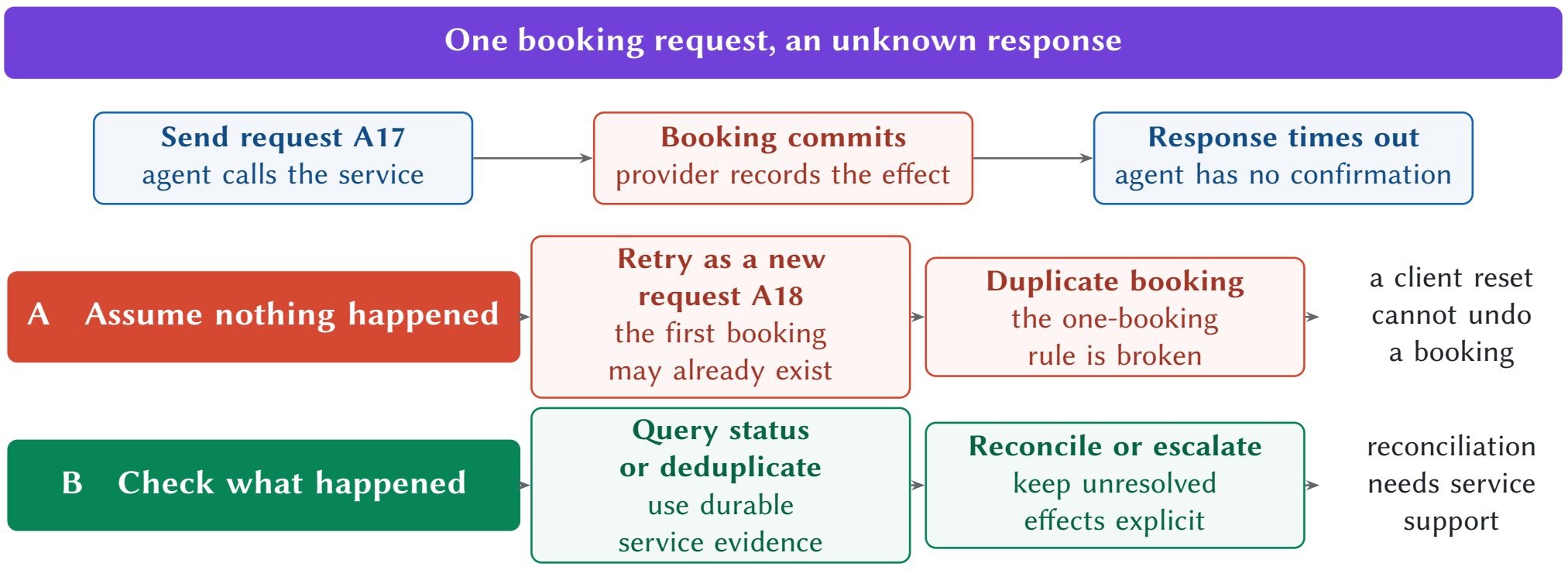

\paragraph{Agent-specific recovery runtimes.} Several 2026 works, mostly preprints, show that checkpoint-and-restore can replay or fork externally visible actions unless it records which effects were released. After a restore, an agent may reconstruct a slightly different request that the server treats as new (Figure~\ref{fig:recoverypaths}; Table~\ref{tab:recoveryruntimes} compares the systems). ACRFence names these \emph{semantic rollback attacks}, including action replay and authority resurrection, and enforces replay-or-fork semantics over recorded irreversible effects~\citep{zheng2026acrfence}. Wu et al.\ identify five checkpoint-and-rollback failure modes across five frameworks, demonstrating malware-verification bypass, unauthorized mail forwarding, and double payment~\citep{wu2026saferesume}. Among constructive systems, Cordon stages outward effects in an outbox and validates the composed flow before release~\citep{chen2026cordon}. Zheng et al.\ decide exactly, with mechanized proofs, whether a checkpoint, fork, restore, or merge is safe~\citep{zheng2026executionedits}. AgentRewind checkpoints agent context with a controlled environment~\citep{zhuang2026agentrewind}, and REVISE recomputes only work invalidated by a revision~\citep{qi2026revise}. Planarian snapshots local state and records compensating actions for remote changes, which rollback replays~\citep{guo2026planarian}, and a commit-time authority service rejects stale proposals and dispatches effects through an idempotent outbox~\citep{salas2026committime}. Recovery can also harm: the same intervention rescues some trajectories and disrupts others, so a router that decides when to intervene raises success~\citep{xiao2026harnesssignal}. All still depend on provider semantics for released effects.

\begin{table}[tbp]
\centering\footnotesize\hyphenpenalty=10000\exhyphenpenalty=10000
\caption{Agent-specific recovery runtimes and studies (2026 preprints). Each makes the recovery boundary explicit; none removes the need to reconcile effects already released to external services.}
\label{tab:recoveryruntimes}
\begin{tabularx}{\linewidth}{@{}P{.17\linewidth}P{.18\linewidth}P{.26\linewidth}P{.13\linewidth}X@{}}
\toprule
\textbf{System}&\textbf{Recovery boundary}&\textbf{Treatment of released effects}&\textbf{Assurance}&\textbf{Residual dependency}\\\midrule
ACRFence~\citep{zheng2026acrfence}&Irreversible tool effects across restore&Effect log; replay or fork on restoration&Attack demonstrations; prototype&Effects must be identified as irreversible\\
Cordon~\citep{chen2026cordon}&Task-level semantic transaction&Shadow state; staged outbox; validation before commit&Experiments&Validation predicates; approval for released effects\\
Execution edits~\citep{zheng2026executionedits}&Checkpoint, fork, restore, merge&Execution record decides which past authorizations and requests an edit must account for&Exact check; mechanized proofs&Complete execution record and policy model\\
AgentRewind~\citep{zhuang2026agentrewind}&Agent context and controlled environment&Aligned checkpoints; resume with prior-attempt information&Experiments&Snapshot-capable environment\\
REVISE~\citep{qi2026revise}&Partially executed workflow&Dependency-guided invalidation and recomputation&Experiments&Recorded data and control dependencies\\
Planarian~\citep{guo2026planarian}&Statepoint over local and remote state&Local snapshot; recorded compensating actions replayed on rollback&Experiments&Compensations exist and succeed\\
Authority at commit time~\citep{salas2026committime}&Admission of a proposal to a governed effect&Idempotent outbox; stale proposals rejected and rerun&Protocol design&Authoritative service; declared dependencies\\
Safe to Resume?\ \citep{wu2026saferesume}&Checkpoint and rollback in five frameworks&Study: five failure modes; demonstrated bypass, forwarding, double payment&Security analysis&Not a mechanism\\
\bottomrule
\end{tabularx}
\end{table}

\subsection{Verification: tests, monitors, and model judges}
\label{sec:mech4}

This subsection covers the checks that decide whether work is acceptable: tests, monitors, and model judges. Runtime verification checks properties against execution events~\citep{leucker2009runtime}, so its verdicts cover only those properties and events. Software tasks offer tests~\citep{jimenez2024swebench,zhang2024autocoderover}, but suites differ in coverage, and stronger tests change which solutions pass~\citep{liu2024evalplus}. WebArena, AppWorld, and $\tau$-bench check environment state~\citep{zhou2024webarena,trivedi2024appworld,yao2024taubench} with privileges deployments may lack, so a benchmark oracle is not a runtime verifier.

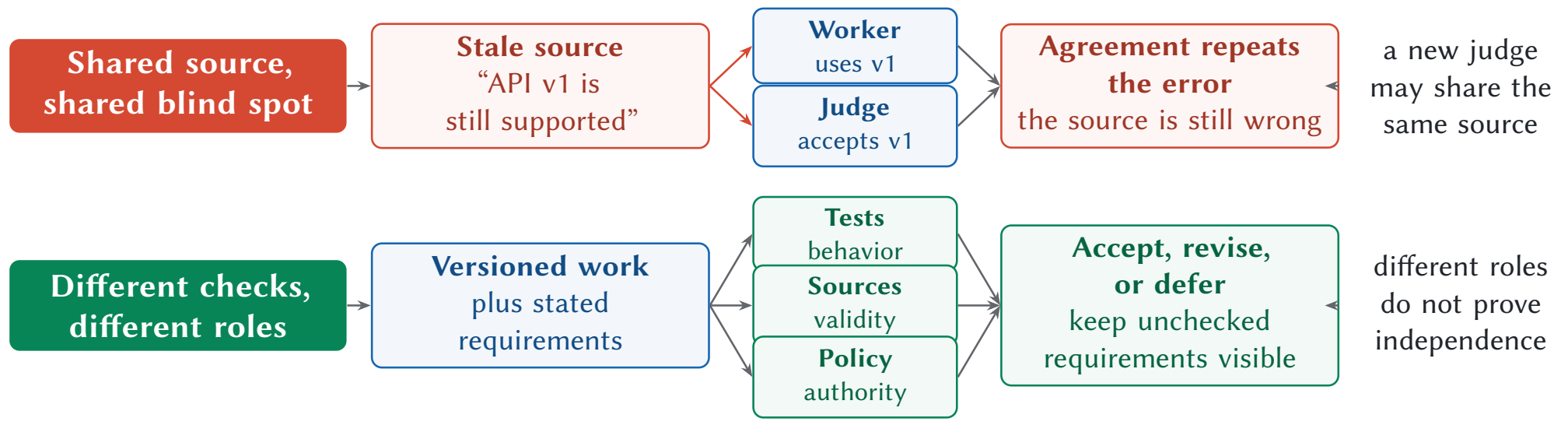
\begin{figure}[tbp]
\centering\resizebox{.98\linewidth}{!}{\definecolor{smA}{HTML}{7040D8}\definecolor{smB}{HTML}{1766B1}\definecolor{smC}{HTML}{078557}\definecolor{smD}{HTML}{D54832}\definecolor{smInk}{HTML}{20242A}\definecolor{smG}{HTML}{6B6F76}
\begin{tikzpicture}[x=1cm,y=1cm,font=\sffamily\small,
 hd/.style={rounded corners=3pt,fill=#1,text=white,font=\sffamily\normalsize\bfseries,text width=3.3cm,minimum height=.95cm,align=center},
 bx/.style={draw=#1,fill=#1!5,rounded corners=3pt,line width=.6pt,align=center,text width=3.3cm,minimum height=.95cm,text=#1!75!black},
 sm/.style={draw=#1,fill=#1!5,rounded corners=3pt,line width=.6pt,align=center,text width=1.9cm,minimum height=.62cm,text=#1!75!black,font=\sffamily\footnotesize},
 cv/.style={align=center,text width=2.6cm,text=smInk},
 ar/.style={-{Stealth[length=4pt]},line width=.6pt,draw=smInk!70}]
\node[hd=smD] (h1) at (1.8,0) {Shared source,\\shared blind spot};
\node[bx=smD] (src) at (5.6,0) {\textbf{Stale source}\\``API v1 is\\still supported''};
\node[sm=smB] (w) at (8.9,0.42) {\textbf{Worker}\\uses v1};
\node[sm=smB] (j) at (8.9,-0.42) {\textbf{Judge}\\accepts v1};
\node[bx=smD] (agr) at (12.2,0) {\textbf{Agreement repeats}\\\textbf{the error}\\the source is still wrong};
\node[cv] (c1) at (15.25,0) {a new judge may share the same source};
\draw[ar] (h1) -- (src);
\draw[ar,draw=smD] (src.east) -- (w.west); \draw[ar,draw=smD] (src.east) -- (j.west);
\draw[ar] (w.east) -- (agr.west); \draw[ar] (j.east) -- (agr.west);
\draw[ar,dashed] (agr) -- (c1);
\node[hd=smC] (h2) at (1.8,-2.3) {Different checks,\\different roles};
\node[bx=smB] (vw) at (5.6,-2.3) {\textbf{Versioned work}\\plus stated\\requirements};
\node[sm=smC] (t) at (8.9,-1.55) {\textbf{Tests}\\behavior};
\node[sm=smC] (s) at (8.9,-2.3) {\textbf{Sources}\\validity};
\node[sm=smC] (p) at (8.9,-3.05) {\textbf{Policy}\\authority};
\node[bx=smC] (acc) at (12.2,-2.3) {\textbf{Accept, revise,}\\\textbf{or defer}\\keep unchecked requirements visible};
\node[cv] (c2) at (15.25,-2.3) {different roles do not prove independence};
\draw[ar] (h2) -- (vw);
\foreach \n in {t,s,p} {\draw[ar] (vw.east) -- (\n.west); \draw[ar] (\n.east) -- (acc.west);}
\draw[ar,dashed] (acc) -- (c2);
\end{tikzpicture}}
\caption{Evidence dependence. Left: worker and judge agree because both rely on one stale source. Right: tests, source checks, and policy checks cover different obligations.}
\label{fig:verification}
\Description{A stale source feeds both worker and judge on the left. On the right, versioned work is checked against behavioral, source, and authority obligations before acceptance or deferral.}
\end{figure}

Model judges assess semantic requirements, and chain-of-verification, debate, and multi-judge systems organize this checking differently~\citep{dhuliawala2024chainverif,du2023debate,chan2024chateval}. MT-Bench and FairEval document judge biases, including sensitivity to answer order~\citep{zheng2023mtbench,wang2024judgebias}. Agent-as-a-Judge inspects intermediate artifacts. On 55 development tasks it outperformed LLM-as-a-Judge and matched the reliability of its human-evaluation baseline~\citep{zhuge2024agentjudge}. Let $W$ denote a wrong worker output and $V$ acceptance, so false acceptance has probability $\Pr(W)\Pr(V\mid W)$. The conditional rate $\Pr(V\mid W)$ records how often the verifier accepts the worker's actual errors. It should be measured because overall accuracy can hide a high value. Deferral thresholds likewise need calibration on the agent's outputs~\citep{guo2017guocal}. A judge adds most value when it reads evidence the worker did not use. A different-model judge helps only if its errors differ from the worker's; error overlap should be measured directly (Figure~\ref{fig:verification}). For approval monitors, every reliable fooling in one study came through a shared evidence source, so counting independent sources rather than votes blocked 0.9\% of correct payments against 7.4\% for the best vote-counting rule at equal safety~\citep{zanolini2026correlatedmonitors}.

\subsection{Coordination protocols and redundancy}
\label{sec:mech5}

This subsection covers how shared work among agents is organized and when redundancy helps. Belief--desire--intention (BDI) models, blackboards, and protocols organize shared work~\citep{wooldridge2009mas,rao1995bdi}, whereas LLM frameworks use role-based pipelines, multi-agent conversations, and state machines~\citep{hong2024metagpt,qian2024chatdev,wu2023autogen,wu2024stateflow}. For example, HuggingGPT and LLMCompiler dispatch specialized models or parallel calls~\citep{shen2023hugginggpt,kim2024llmcompiler}. Role names do not supply dependability, which requires enforced ownership, versioned handoffs, restricted writes, and bounded termination. Self-consistency, debate, and mixtures of agents aid candidate selection~\citep{wang2023selfconsistency,du2023debate,wang2024mixtureofagents}, but, like N-version programming and recovery blocks~\citep{avizienis1985nversion,randell1975faulttol,lyu1995softwareft}, they depend on diversity, because replicas of one model tend to share errors. Agreement protocols~\citep{schneider1990smr,castro1999pbft} make replicas agree but not correct, a limit that majority voting among agents shares.

\subsection{Observability, human review, and audit}
\label{sec:mech6}

This subsection covers the records and human processes that support review during and after a run. Records serve three purposes: tracing, review, and audit. Distributed tracing links events through causal context~\citep{sigelman2010dapper}, but agents also need semantic links, such as links from sources to claims and from approvals to effects. Who\&When, AgentRx, LongRCA, and TRAIL localize responsible agents, steps, and error types~\citep{zhang2025whowhen,barke2026agentrx,zhang2026longrca,deshpande2025trail}. Dependency-guided attribution outperforms chronology-based judges on Who\&When~\citep{deng2026defa}, and BRIDGE compiles handoff traces into a provenance graph to repair broken handoffs at run time~\citep{su2026bridge}. The Agent Error Dataset pairs 50,000 failures with replay-checked diagnoses~\citep{zhu2026aed}, while an audit of attribution benchmarks finds surface cues that let simple predictors match GPT-4.1~\citep{xu2026auditattribution}. Review packets should therefore expose proposed effects, checks, uncertainty, and required decisions. READY selects the minimum-cost oversight policy meeting a reliability target and qualifies it statistically~\citep{chatrath2026ready}, LEASH escalates from a weak to a strong computer-use agent when step-level uncertainty rises~\citep{wang2026leash}, while cognitive-forcing experiments show that interface design changes overreliance~\citep{bucc2021cognitiveforcing}. Model cards and datasheets are precedents for agent records~\citep{mitchell2019modelcards,gebru2021datasheets}, and such records need maintenance as systems change~\citep{sculley2015hidden}.

\subsection{Composition: align control and effect boundaries}
\label{sec:mech7}

Combining controls is safe only if the state each component checks is the state the next uses. We represent a control by $(A_i,G_i,\Omega_i,\mathcal{B}_i)$: assumptions, guarantee, observable state, and enforcement or restoration boundary. In Table~\ref{tab:controlcontracts}, $A_i$, $\Omega_i$, and $\mathcal{B}_i$ correspond to the columns \emph{principal assumptions}, \emph{inspection basis}, and \emph{enforces or restores}. The \emph{protects} column names the state domain that $G_i$ is about, whereas $G_i$ itself is a property of that state. A test gate, for example, protects the repository (R4), and its guarantee is that commit $v$ passes test suite $T$; a release step that follows assumes that the artifact it deploys is $v$. Cost and residual risk are not part of the contract. For consecutive deterministic controls under environment assumptions $\Gamma$, a sufficient compatibility obligation is
\begin{equation}
\Gamma\land G_i\ \models\ A_{i+1}.
\label{eq:contract}
\end{equation}
A gap can remain between check and commit, for example between tested version $v$ and release $v+1$, or approved arguments $a$ and rewritten $a'$. Where supported, version binding or atomic compare-and-update closes this gap.

For two probabilistic checks with miss events $M_1$ and $M_2$, the joint miss probability is $\Pr(M_1)\Pr(M_2\mid M_1)$. The second check's value therefore depends on its miss rate \emph{on the first check's misses}, not its marginal accuracy. Checks reading the same inputs can be expected to have $\Pr(M_2\mid M_1)$ well above $\Pr(M_2)$, so this dependence should be measured rather than assumed.

Recovery must also leave every running component consistent; restoring a database while a client keeps its old view violates this. Table~\ref{tab:controlcontracts} compares the control families of Sections~\ref{sec:mech1}--\ref{sec:mech6} along six dimensions.

\begin{table}[tbp]
\centering\scriptsize
\setlength{\tabcolsep}{2.5pt}
\renewcommand{\arraystretch}{1.12}
\caption{Comparison of control families. \emph{Inspection basis}: what the control observes; \emph{Protects}: state domains whose integrity it targets; \emph{Enforces or restores}: what it can block before an effect or return to a supported state, if anything. The table compares contracts, not measured efficacy.}
\label{tab:controlcontracts}
\begin{tabularx}{\linewidth}{@{}P{.16\linewidth}P{.14\linewidth}P{.06\linewidth}P{.11\linewidth}P{.17\linewidth}P{.1\linewidth}X@{}}
\toprule\textbf{Family (examples)}&\textbf{Inspection basis}&\textbf{Protects}&\textbf{Enforces or restores}&\textbf{Principal assumptions}&\textbf{Added cost}&\textbf{Residual risk}\\\midrule
\multicolumn{7}{@{}l}{\textit{Consistent decisions (Sections~\ref{sec:mech1}--\ref{sec:mech2})}}\\
Search, decomposition~\citep{yao2023tot,zhou2023lats,meyerson2025million}&Candidate paths scored by an evaluator or local votes&R2&Backtrack within the plan&Evaluator discriminates; subtasks separable&Multiples of model calls&Shared evaluator error; effects of explored branches unless sandboxed\\
Reflection, revision~\citep{shinn2023reflexion,gou2024critic}&Own outputs, feedback, tool results&R2&A further attempt, not state&Feedback is informative about the error&Extra trials&Reinforcing a wrong diagnosis\\
Environment-verified task state~\citep{ma2026longhorizonharness,wang2024voyager}&Observation before a belief or skill is stored&R2 vs R4&Re-derive the belief&Relevant state is observable and cheap to read&Read and audit calls&Unobservable or lagging state\\
Context compression, KV retention~\citep{kang2026acon,zhang2023h2o,li2024snapkv}&Attention scores or learned compression rules&R1, R2&Nothing (lossy)&Importance tracks later needs&Saves resources&Silent loss of a constraint\\
Retrieval, memory with provenance~\citep{asai2024selfrag,yan2024crag,chhikara2025memzero}&Relevance, recency, update operations&R2, R5&Update or delete items&Provenance recorded; invalidation reaches consumers&Storage, retrieval&Stale or poisoned items; dependants not invalidated\\\midrule
\multicolumn{7}{@{}l}{\textit{Limiting incorrect effects (Sections~\ref{sec:mech3}--\ref{sec:mech4})}}\\
Schemas, constrained decoding~\citep{willard2023grammar,beurerkellner2023lmql}&Token stream&R3&Nothing&Syntax is the property of interest&Low&Valid but wrong calls\\
Deterministic runtime policy~\citep{agentspec2025,shi2025progent}&Each call and its arguments&R3, R4, R6&Blocks before effect&Complete mediation; adequate rules&Policy authoring&Uncovered hazards; over-blocking\\
Information-flow separation~\citep{debenedetti2025camel}&Provenance of data and control flow&R2, R3&Blocks before effect&Program model captures flows; trusted planner&Utility loss on some tasks&Tasks needing data-dependent control; uncovered channels\\
Learned guards, trained robustness~\citep{inan2023llamaguard,xiang2025guardagent,chen2025shieldagent,wallace2024hierarchy}&Text or proposed actions; implicit in the model&R2, R3, R6&Blocks or flags; nothing if implicit&Training or policy extraction matches deployment&Extra model call or training&Measured false acceptance; adaptive attacks; no audit record if implicit\\
Isolation, emulation~\citep{agache2020firecracker,ruan2024toolemu}&Reachable resources&R4&Discard the sandbox&Effects confined to the sandbox&Setup&Simulation gap; allowed channels\\
Checkpoint and restore~\citep{elnozahy2002rollback,zhuang2026agentrewind}&Snapshot of context and controlled environment&R2, R4&Restore the snapshot&Snapshot covers dependent state&Storage, latency&External effects; concurrent readers\\
Transactional staging~\citep{chen2026cordon}&Composed outbound effects before commit&R4, R6&Discard staged effects&Effects deferrable; sound validation predicates&Latency, approval&Effects released earlier; predicate gaps\\
Effect logs, replay-or-fork~\citep{zheng2026acrfence,zheng2026executionedits}&Record of irreversible effects&R2, R4&Replay recorded result or fork&Irreversible effects identified; complete record&Logging&Provider semantics; unlogged effects\\
Tests, state oracles~\citep{jimenez2024swebench,trivedi2024appworld}&Artifact or environment state against a predicate&R4&Nothing (detects)&An oracle exists and covers the requirement&Execution&Untested requirements; benchmark privileges\\
Model judges~\citep{zheng2023mtbench,zhuge2024agentjudge}&Outputs, or artifacts and traces&R2, R4&Nothing (detects)&Independent evidence; calibrated on worker errors&Model calls&Correlated misses; position bias\\\midrule
\multicolumn{7}{@{}l}{\textit{Governing shared work (Sections~\ref{sec:mech5}--\ref{sec:mech6})}}\\
Redundancy, voting, debate~\citep{wang2023selfconsistency,du2023debate,wang2024mixtureofagents}&Several candidates&R2, R5&Choose a candidate&Error diversity&Multiples of model calls&Common-mode errors\\
Coordination protocols~\citep{wu2024stateflow,hong2024metagpt}&Message and state transitions&R5&Re-enter a protocol state&Protocol enforced, not only prompted&Design&Unenforced roles; stale reads\\
Human review, audit~\citep{chatrath2026ready,bucc2021cognitiveforcing}&Review packet of effects and evidence&R6&Veto, escalate&Attention, evidence, and authority available&Human time&Over-reliance; latency on long traces\\
\bottomrule
\end{tabularx}

\end{table}

\paragraph{Choosing among controls.} The choice of control should follow the error's boundary. Hazards expressible as predicates over actions or information flow suit deterministic enforcement, which holds by construction within its mediation; semantic hazards suit learned guards whose false-acceptance rate is measurable on the deployment distribution. Where effects can be deferred, staging removes the need to undo them. For effects that cannot be deferred, the remaining defenses are effect logs with replay-or-fork and provider-side idempotence, since snapshot restore alone suffices only for snapshot-capable state. Where an executable oracle exists, tests dominate model judges for the requirements they cover; elsewhere, judges help only when they read evidence the worker did not use, and redundancy only when error diversity is shown.

\section{Evaluation: Outcomes, Episodes, and Causal Claims}
\label{sec:evaluation}

\begin{figure}[tbp]
\centering\resizebox{.98\linewidth}{!}{\definecolor{smA}{HTML}{7040D8}\definecolor{smB}{HTML}{1766B1}\definecolor{smC}{HTML}{078557}\definecolor{smD}{HTML}{D54832}\definecolor{smInk}{HTML}{20242A}
\begin{tikzpicture}[x=1cm,y=1cm,font=\sffamily\small,
 hd/.style={rounded corners=3pt,fill=#1,text=white,font=\sffamily\normalsize\bfseries,minimum width=3.3cm,minimum height=.95cm,align=center},
 bx/.style={draw=#1,fill=#1!5,rounded corners=3pt,line width=.6pt,align=center,text width=3.9cm,minimum height=.95cm,text=#1!75!black},
 cv/.style={align=center,text width=3.6cm,text=smInk,font=\sffamily\small},
 ar/.style={-{Stealth[length=4pt]},line width=.6pt,draw=smInk!70},
 band/.style={draw=#1,fill=#1!5,rounded corners=3pt,line width=.6pt,align=center,minimum width=16.2cm,minimum height=.75cm,text=smInk}]
\node[band=smA,fill=smA,text=white,font=\sffamily\normalsize\bfseries] at (8.1,0) {Choose the population before choosing the score};
\foreach \y/\c/\h/\p/\m/\v in {
 -1.3/smB/Task evaluation/{\textbf{Selected tasks}\\keep successes and failures}/{\textbf{Measure}\\completion, impact, cost}/{may not represent deployment},
 -2.6/smC/Failure diagnosis/{\textbf{Failed trajectories}\\keep label uncertainty}/{\textbf{Measure}\\fault location, repair value}/{does not measure fault frequency},
 -3.9/smD/Controlled stress/{\textbf{Attacks or faults}\\keep clean controls}/{\textbf{Measure}\\containment and utility}/{does not measure natural incidence}}{
 \node[hd=\c] (h) at (1.65,\y) {\h};
 \node[bx=\c] (a) at (5.75,\y) {\p};
 \node[bx=\c] (b) at (10.6,\y) {\m};
 \node[cv] (c) at (14.35,\y) {\v};
 \draw[ar] (h) -- (a); \draw[ar] (a) -- (b); \draw[ar,dashed] (b) -- (c);}
\node[band=smC] at (8.1,-5.2) {\textbf{In every design:} keep versions, budgets, denominators, and outcomes};
\end{tikzpicture}}
\caption{Evaluation begins with population selection: task cohorts, failure-conditioned corpora, and constructed stress tests support different claims, and their denominators remain separate.}
\label{fig:evaluationmap}
\Description{Three evaluation lanes show task cohorts, failed-trace diagnosis, and adversarial stress, each linked to its measures and inference limits.}
\end{figure}
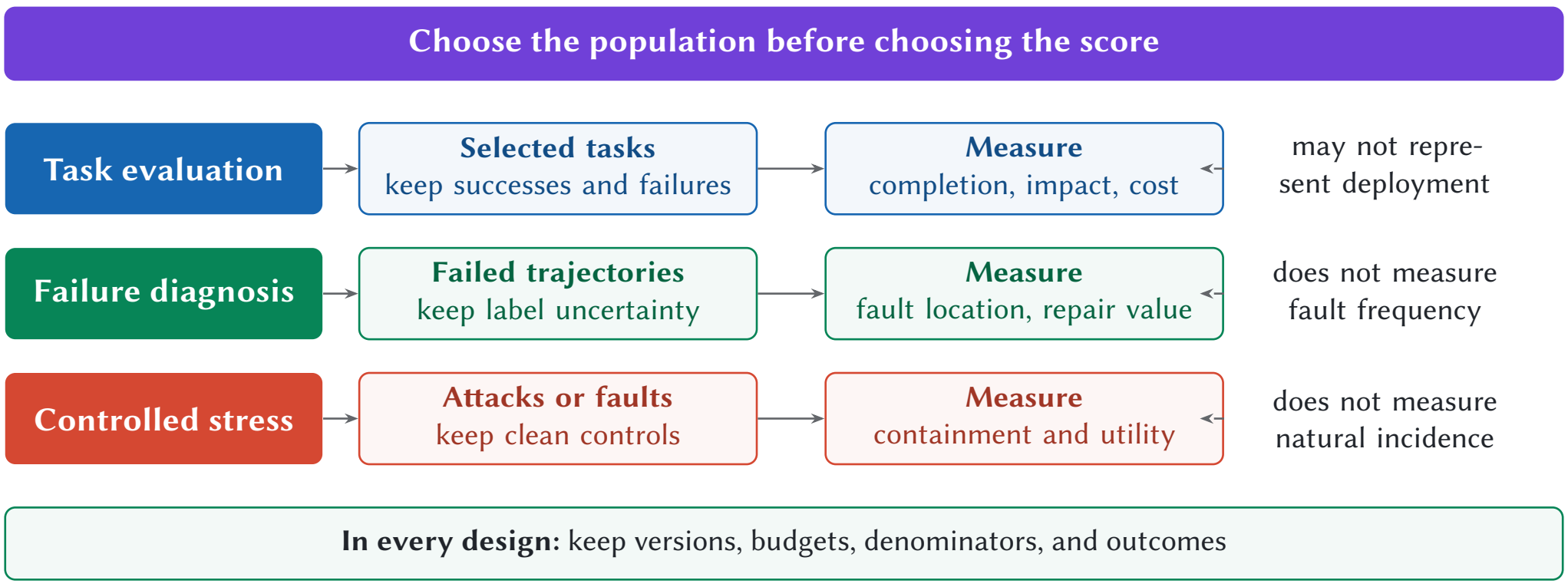

The value of a benchmark score depends on what the test observes and which runs it includes, so task cohorts, failure-conditioned corpora and constructed stress tests support different claims even with the same model and environment (Figure~\ref{fig:evaluationmap}). We compare task benchmarks (Section~\ref{sec:families}) with adversarial and diagnostic studies (Section~\ref{sec:diagnostic}) and propose measurements and designs (Sections~\ref{sec:measures}--\ref{sec:bmiss}).

\subsection{Task benchmarks and their observability}
\label{sec:families}

SWE-bench and SWE-Bench Pro test repository-scale correctness via versioned tasks and executable tests~\citep{jimenez2024swebench,deng2025swebenchpro}, so results depend on test adequacy, environment setup, and release version. WebArena, OSWorld, and AppWorld expose web, desktop, and API actions with functional or state-based evaluators~\citep{zhou2024webarena,xie2024osworld,trivedi2024appworld}, whereas Mind2Web's offline task measures reproduction of demonstrated actions~\citep{deng2023mind2web}. Further benchmarks add visual grounding, enterprise software, mobile, conversational navigation, and lengthy open-web tasks~\citep{koh2024visualwebarena,drouin2024workarena,rawles2025androidworld,l2024weblinx,yoran2024assistantbench}. Others cover heterogeneous environments, progress, API use, and tool-supported answers~\citep{liu2024agentbench,ma2024agentboard,qin2024toolbench,mialon2023gaia}, with BrowserGym unifying evaluation~\citep{chezelles2025browsergym}. $\tau$-bench scores database state under policies with simulated users~\citep{yao2024taubench}. In $\tau^2$-bench, agent and user both act on shared state, so evaluation must check its consistency~\citep{barres2026tau2bench}.

Newer benchmarks target consequential, long-running work: a simulated software company~\citep{xu2024theagentcompany}, Kaggle competitions~\citep{chan2025mlebench}, priced freelance tasks~\citep{miserendino2025swelancer}, hard command-line tasks~\citep{merrill2026terminalbench}, and 20-million-token business simulations~\citep{backlund2025vendingbench}. RE-Bench, comparing agents with eight-hour expert attempts~\citep{wijk2024rebench}, shows that budget length decides who looks stronger. Agents scored four times higher than humans at two hours; humans scored twice as high at 32 total hours. Because these benchmarks stress different components of the horizon (Section~\ref{sec:traj}) and measure horizon differently, their scores share no single horizon scale.

\subsection{Adversarial, long-horizon, and diagnostic evidence}
\label{sec:diagnostic}

Security benchmarks each define an attacker surface and threat model. AgentDojo pairs legitimate tasks with injections, while InjecAgent injects malicious tool content~\citep{debenedetti2024agentdojo,shi2024injecagent}. Agent Security Bench broadens attack and defense settings, whereas AgentHarm tests explicitly harmful requests~\citep{zhang2025asb,andriushchenko2025agentharm}. Indirect hijacking and deliberate misuse are distinct threats, so report their rates separately. Long-horizon stress tests reach hundreds of tool calls or thousands of trajectories~\citep{luo2025ultrahorizon,wang2026horizon}. Recovery now has benchmarks of its own: CUA-Recovery places agents in 2,250 verified erroneous desktop states, where error awareness reaches at most 44\%~\citep{hu2026cuarecovery}; ReTrace judges a recovery by whether downstream actions yield the same outcomes as from the target state~\citep{bang2026lostrecovery}; and UndoBench and LIMBO score recovery against ledgers of committed effects~\citep{sah2026undobench,li2026limbo}. Memory benchmarks isolate temporal reasoning, updates, abstention, and selective forgetting~\citep{maharana2024locomo,wu2025longmemeval,hu2026memoryagentbench}.

Failure-attribution corpora identify which role or step explains a failed run~\citep{zhang2025whowhen,deshpande2025trail,barke2026agentrx,zhang2026longrca}. Step-level diagnosis is far harder. The best Who\&When method found the responsible agent 53.5\% of the time, but the decisive step only 14.2\%. In LongRCA, root causes lie a median 48 steps before run end, and the strongest baseline reached 13.2\% exact root-step accuracy. TRAVERSE adds 2,518 human-annotated trajectories, yet the best of six frontier judges found the first mistake in under a third of runs~\citep{rahman2026traverse}. These corpora assess diagnosis of known failures only, not failure frequency or whether one root step exists, and their labels have their own uncertainty: MAST's $\kappa=0.88$ was measured by three annotators on 15 of its 150 development traces, and its 1,642-trace dataset is labeled by an LLM judge~\citep{cemri2025mast}. Role accuracy, step accuracy, and repair usefulness therefore need separate reporting~\citep{wang2026trajectorysurvey,chen2026trajectoryattribution}.

\subsection{A minimal measurement protocol}
\label{sec:measures}

Trajectory-level claims need a minimal record so that outcome changes can be attributed to the right harness change: task, trial, versions, budget, realized horizon, disposition, and harness configuration (model, prompts, tools, memory, verifier, recovery policy, human boundary).

Repeated-trial reliability needs care because common metrics answer different questions. \emph{Pass@$k$} asks whether at least one of $k$ trials succeeds, whereas $\mathrm{pass}^{k}$, introduced in $\tau$-bench~\citep{yao2024taubench}, asks whether all $k$ succeed. For $n$ reset trials with $c$ successes, and assuming exchangeable trials (valid when earlier attempts leave external state unchanged for later trials),
\begin{equation}
\widehat{\mathrm{pass@}k}=1-\frac{\binom{n-c}{k}}{\binom{n}{k}},
\qquad
\widehat{\mathrm{pass}^{k}}=\frac{\binom{c}{k}}{\binom{n}{k}},\quad k\leq n.
\label{eq:pass}
\end{equation}

Verifier coverage also needs a denominator, which is the set of applicable requirement--event obligations $\mathcal{O}(\tau)$ with checking deadlines:
\begin{equation}
\operatorname{Cov}(\tau)=
\frac{|\{o\in\mathcal{O}(\tau):\text{the required check ran before its deadline}\}|}
{|\mathcal{O}(\tau)|}.
\label{eq:coverage}
\end{equation}
It counts executed checks, not their soundness. Reporting it by requirement type prevents repeated easy checks from hiding an untested one. Cost per success, $\sum_i C_i/\sum_iY_i$, must include failed attempts, recovery, and review effort~\citep{chatrath2026ready}.

\subsection{Experimental designs that isolate a mechanism}
\label{sec:horizoncontrol}

Three designs each change one preset property while keeping the task goal fixed. The first two vary horizon and dependency structure, so the compared tasks must match on everything except duration or dependency distance. The third studies error propagation and containment.
A \emph{dependency-delay intervention} varies the steps between observing a fact and using it, crossed with the summarization policy; a \emph{repeated-operation intervention} adds validated operations whose postconditions are known; and a \emph{state intervention} injects a false premise, stale response, omitted message, or numerical perturbation at a declared phase and domain, which tests containment under that injection. Each design should randomize at task or trial level, retain clean controls, include successful and failed runs, and compare at a fixed budget and at equal task success. Crossing two controls factorially estimates their interaction, which tests whether aligned controls reduce escaping errors.

\subsection{Statistical reporting}
\label{sec:bmiss}

Estimate uncertainty by treating tasks as clusters, because steps correlate, and audit a subset of model-judged outcomes for false acceptance and rejection. Under a fixed budget, count every attempted task and distinguish completion, policy failure, escalation, and timeout. For time-to-failure estimates, treat successful completion as a competing terminal event and report cumulative incidence; only stops unrelated to failure risk, such as a budget cutoff, count as censored. Combining failure-enriched corpora with representative runs requires a sampling model and weights, while stratifying by realized length requires addressing errors that lengthen runs.

\section{Published Evidence and Research Priorities}
\label{sec:open}

\subsection{Findings supported by published results}
\label{sec:findings}

We grade each finding's evidence: \emph{strong by construction} if it rests on a demonstrated mechanism or a design guarantee, \emph{moderate} if consistent but observational or setting-specific, and \emph{descriptive} if it concerns research attention rather than system behavior. F1--F3 concern where errors enter and spread, F4--F6 which controls interrupt them, and F7 and OP1--OP5 what evaluations establish.
\begin{itemize}
\item \textbf{F1. Errors make later errors more likely} (moderate). In coding, tool-use, and controlled reasoning tasks, error rates after an error exceed baseline, and failure onset tends to precede symptoms by many steps~\citep{sinha2025illusion,lee2026canonical,yang2024sweagent,zhao2026failureprocess,rahman2026traverse}. A history-dependent hazard (Equation~\ref{eq:hazard}) is therefore a better starting model than independent steps. Outside injected-history and natural-experiment designs, part of the effect may reflect task difficulty.
\item \textbf{F2. Length alone does not explain reliability} (moderate). Decomposition with local error control sustained a million-step task. Vending-Bench derailment decouples from context saturation, and the RE-Bench human--agent ranking reverses with the time budget~\citep{meyerson2025million,backlund2025vendingbench,wijk2024rebench}. Claims about ``long-horizon failure'' should therefore name the changed horizon component and dependency structure.
\item \textbf{F3. Losing evidence and failing to use it are different faults} (moderate). Long-context studies show evidence present but unused, and planning agents leave gathered constraints unused~\citep{liu2024lostmiddle,xie2024travelplanner}. Memory and compression studies instead show evidence lost before use~\citep{hu2026memoryagentbench}. Compression should thus be judged by its effect on later decisions~\citep{kang2026acon}.
\item \textbf{F4. For indirect prompt injection, data--authority separation is a central security boundary} (strong by construction). Attacks through content, memory, or tool metadata succeed when data gains the power to trigger actions~\citep{greshake2023greshake,debenedetti2024agentdojo,hou2025mcp}. Architectural separation with deterministic privilege control blocks these attacks by design when the policy is correct and covers the relevant operations~\citep{debenedetti2025camel,shi2025progent}, a guarantee different in kind from the probabilistic robustness of trained defenses, although it does not compose across agents that call one another~\citep{petersgill2026camelstalk}. The utility cost is measurable: on AgentDojo, CaMeL completes 77\% of tasks against 84\% without a defense~\citep{debenedetti2025camel}.
\item \textbf{F5. Local checkpointing alone does not make retry of external actions safe} (strong by construction; defenses are recent preprints). Restored agents re-issue already committed actions, including payments~\citep{wu2026saferesume,zheng2026acrfence}, and recovery benchmarks measure duplicate effects in half or more of naive retries~\citep{li2026limbo,sah2026undobench}. Safe recovery needs a record of released effects and an explicit restoration boundary~\citep{chen2026cordon,zheng2026executionedits}.
\item \textbf{F6. Agreement is weak evidence when evidence is shared} (moderate). Self-correction can degrade reasoning, judges show position bias, and on TRAIL the best model localizes only 11\% of trace errors~\citep{huang2024selfcorrect,zheng2023mtbench,wang2024judgebias,deshpande2025trail}. Gains appear when tests or inspected artifacts supply new evidence~\citep{zhuge2024agentjudge}. A verifier therefore adds value through independent evidence or measured error diversity, and approval rules should count independent sources rather than votes~\citep{zanolini2026correlatedmonitors}.
\item \textbf{F7. Benchmark success and deployability differ} (moderate). Repeated-trial reliability falls far below single-trial success. On $\tau$-bench retail, $\mathrm{pass}^8$ was below 25\% for a model succeeding on under half of tasks~\citep{yao2024taubench}. Final-state checks miss collateral changes unless they test for them, as AppWorld does~\citep{trivedi2024appworld}, so runs scored as solved can still delete data~\citep{rahman2026traverse}. Equal accuracy can demand very different amounts of human review~\citep{chatrath2026ready}.
\end{itemize}

\subsection{Illustrative case analysis: what an episode record adds}
\label{sec:worked}

We test the framework on two published failure traces: the running example from TheAgentCompany~\citep{xu2024theagentcompany} and a multi-agent trace that MAST publishes with human annotations (Appendix N.8 of~\citep{cemri2025mast}). We classify each failure with the symptom label given by its source, with Ding et al.'s state-centered axes~\citep{ding2026alwayson}, and with the episode record (Equation~\ref{eq:episode}). The facts come from the released trajectories; the classification and the indicated controls are ours (Table~\ref{tab:casecompare}).

\paragraph{Case A: a shortcut that changes the environment, then a wait that never ends.} The trace is the released trajectory of the Amazon Nova Pro run on the benchmark's example task~\citep{xu2024theagentcompany}, introduced in Section~\ref{sec:intro}; the paper describes the pattern as renaming another user, and the released trace shows the agent renaming its own administrator account. The agent, logged into the company chat as an administrator, is to obtain a repository location from a colleague, Alex Turner, clone the repository, and start a server. Its second browser action opens the user administration page, and seven actions later it has set the name and username of its own account to ``Alex Turner'', so the directory lists two users of that name. After a guessed repository URL, a guessed wiki URL, and two web searches fail, it opens a direct message with a different colleague, Mike Chen, and writes ``Hi Alex, could you please provide me with the repository URL''. From then on it answers every prompt to continue by saying that it is waiting for Alex Turner, 81 times, until the run ends after 22 minutes without a finish action. The benchmark's authors report this pattern as a fabricated shortcut; the state-centered axes flag an identity record changed without authority. The episode record adds the chain (Table~\ref{tab:casecompare}). As Section~\ref{sec:episode} recorded, the origin is a P1 decision in R2 to alter the directory rather than search it; the P2 action mutates R4 and R6, so an identity in a shared system points at the agent itself; the wrong-recipient message (R5) contaminates the R2 belief at P3 (H2); the directory listing and the message header displayed the evidence, but no P4 check read them, and no P5 response or P6 escalation followed before the budget ran out (H7). Three controls are indicated: a P0 scope rule that excludes directory writes from a task that only reads it, a P2 recipient check against the conversation partner, and a P5 stall rule that escalates after a bounded number of identical waiting turns. A local checkpoint would address none of these, because restoring the agent's state leaves the renamed account in place (F5).

\paragraph{Case B: an example output accepted as a test result.} HyperAgent, a planner--navigator--executor system, works on a Matplotlib issue from SWE-bench Lite in which a RangeSlider callback blocks input to other widgets. The released trace runs through seven planner rounds and six sub-agent handoffs (four navigator, one editor, one executor) with 23 inner tool steps; the executor's report is the last handoff. After the planner's change, the executor writes and runs a test script. The script opens an interactive window, so its only output is a font-cache message. The executor notes that the plot must be operated by hand. Even so, its report to the planner lists interaction steps and concludes that the blocking issue ``has been resolved''. The report ends: ``The above output is just an example''. The planner replies ``Thank you for running the test and providing the results'', states that the solution ``has been tested'', and ends the task~\citep{cemri2025mast}. MAST labels the trace \emph{disobey role specification} (1.2) and \emph{information withholding} (2.4), which describe the navigator's and executor's behavior. The state-centered axes flag a stored claim without provenance. The episode record shows where the evidence changed status. An R4 observation (the font-cache message) and the executor's hypothetical output enter the R5 handoff together. The qualifier does not survive the planner's P3 integration. P4 acceptance and P6 termination therefore rest on example output. In terms of Equation~\ref{eq:coverage}, the trace shows no completed interactive test, yet the run recorded the check as passed. The indicated control binds every ``tested'' claim to an execution record that the planner reads directly; for a GUI fix, the record includes an interaction log or widget-level assertions, not only terminal output. It also blocks termination while a required check is unexecuted. The case also reveals a measurement gap. An end-state score records whether the final patch passes hidden tests, not whether the agent's own verification took place.

\begin{table}[!htbp]
\centering\footnotesize
\setlength{\tabcolsep}{4pt}
\caption{The same failures under three framings. The source paper names the symptom and the state-centered axes identify the vulnerable state item; the episode record adds where information entered, how its status changed, and which control and measurement would catch it.}
\label{tab:casecompare}
\begin{tabularx}{\linewidth}{@{}P{.17\linewidth}XX@{}}
\toprule&\textbf{Case A: shortcut that alters a shared system (TheAgentCompany)}&\textbf{Case B: example accepted as test result (MAST N.8)}\\\midrule
Symptom-level (source)&Fabricated shortcut (``deceiving oneself'')&1.2 disobey role specification; 2.4 information withholding\\
State-centered (Ding et al.)&Identity record changed without authority&Stored ``tested'' claim without provenance\\
Episode: origin&P1/R2: decision to alter the directory instead of searching it&P3/R5: hypothetical output merged with an R4 observation in a handoff\\
Episode: detection, response&Evidence displayed (duplicate name, wrong recipient); no check reads it; no escalation&Executor's caveat present; dropped at integration; P4 accepts\\
Episode: boundary crossed&P2 write $\to$ R4/R6 identity $\to$ R5 message to wrong user $\to$ R2 belief $\to$ 81 waiting turns (H7)&R4 observation and example $\to$ R5 report $\to$ P4 acceptance $\to$ P6 termination\\
Indicated control&Scope rule on directory writes; recipient check; stall rule that escalates&Bind ``tested'' claims to execution records; block termination on unexecuted checks\\
Evaluation implication&Count out-of-scope external mutations and unrecovered stalls alongside end-state success&Report verifier coverage (Equation~\ref{eq:coverage}) with end-state success\\
\bottomrule
\end{tabularx}

\end{table}

The three framings complement one another. A symptom taxonomy is cheap and supports prevalence counts, and a state-centered analysis identifies which items need governance. The episode record needs event order, message provenance, and execution records; in return, it separates failures that share a label but require different controls.

\subsection{Costs and interference}
\label{sec:sideeffects}

Controls add cost and new failure paths: search and voting use compute, provenance uses storage, approval uses attention, compression can lose a constraint, conservative checks can reject legitimate actions, and staging adds latency~\citep{chen2026cordon}. Claiming reward hacking requires evidence that optimization raises acceptance without improving independent correctness~\citep{liu2024evalplus}. Trace-length effects on quality remain unmeasured despite widespread review~\citep{pan2025map}, so compliant completion, unrecovered impact, latency, compute, and human effort should be reported together.

\subsection{Scope and limitations of the background}
\label{sec:limitations}

The findings above rest on published results and inherit their limits: most benchmarks publish end states rather than intermediate state, message provenance, or execution records, so propagation is rarely observable; our labels for what a method restores or contains follow the authors' descriptions; many reported measurements lack the denominators needed to relate them to a population of runs, which is why findings are graded rather than pooled; and software and question-answering tasks supply most of the evidence. Section~\ref{sec:results} addresses the first of these limits directly by annotating released trajectories.

\subsection{Five research priorities}
\label{sec:openproblems}

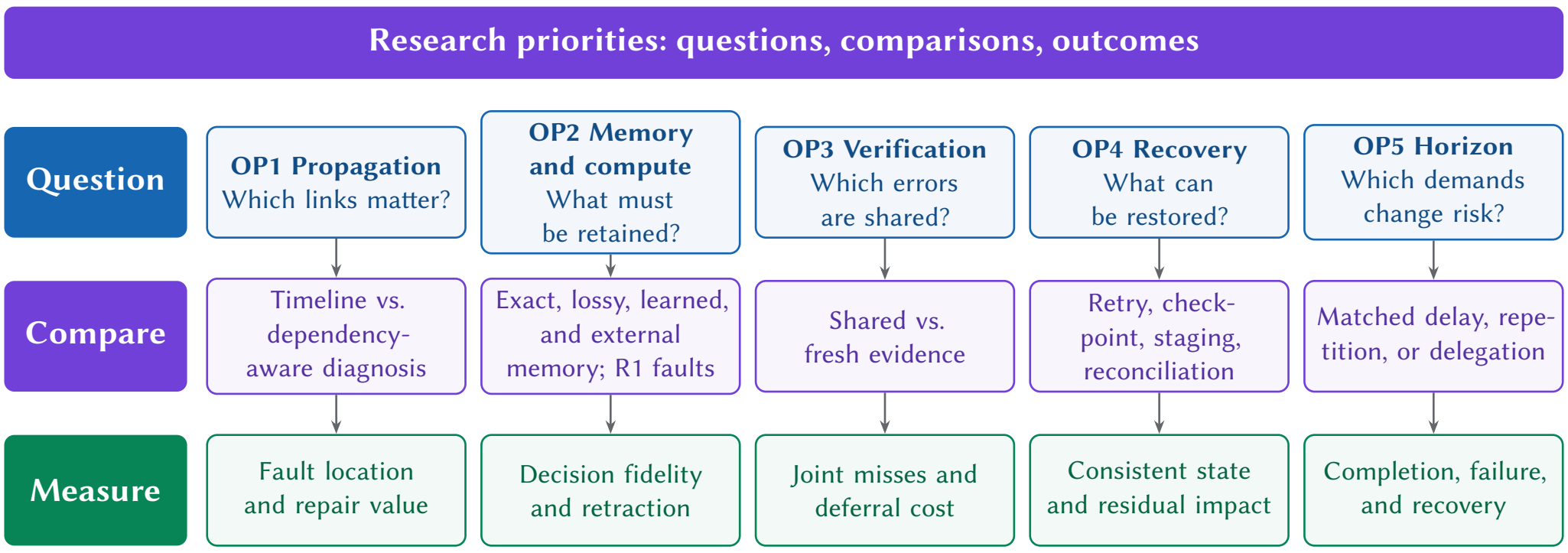
\begin{figure}[tbp]
\centering\resizebox{.98\linewidth}{!}{\definecolor{smA}{HTML}{7040D8}\definecolor{smB}{HTML}{1766B1}\definecolor{smC}{HTML}{078557}\definecolor{smD}{HTML}{D54832}\definecolor{smInk}{HTML}{20242A}
\begin{tikzpicture}[x=1cm,y=1cm,font=\sffamily\small,
 rh/.style={rounded corners=3pt,fill=#1,text=white,font=\sffamily\normalsize\bfseries,minimum width=1.9cm,minimum height=1.15cm,align=center},
 bx/.style={draw=#1,fill=#1!5,rounded corners=3pt,line width=.6pt,align=center,text width=2.45cm,minimum height=1.15cm,text=#1!75!black,font=\sffamily\footnotesize},
 ar/.style={-{Stealth[length=4pt]},line width=.6pt,draw=smInk!70},
 band/.style={rounded corners=3pt,fill=#1,text=white,font=\sffamily\normalsize\bfseries,minimum width=16.2cm,minimum height=.75cm,align=center}]
\node[band=smA] at (8.1,0) {Research priorities: questions, comparisons, outcomes};
\node[rh=smB] at (0.95,-1.45) {Question};
\node[rh=smA] at (0.95,-3.05) {Compare};
\node[rh=smC] at (0.95,-4.65) {Measure};
\foreach \x/\q/\c/\m in {
 3.45/{\textbf{OP1 Propagation}\\Which links matter?}/{Timeline vs.\ dependency-aware diagnosis}/{Fault location and repair value},
 6.3/{\textbf{OP2 Memory and compute}\\What must be retained?}/{Exact, lossy, learned, and external memory; R1 faults}/{Decision fidelity and retraction},
 9.15/{\textbf{OP3 Verification}\\Which errors are shared?}/{Shared vs.\ fresh evidence}/{Joint misses and deferral cost},
 12.0/{\textbf{OP4 Recovery}\\What can be restored?}/{Retry, checkpoint, staging, reconciliation}/{Consistent state and residual impact},
 14.85/{\textbf{OP5 Horizon}\\Which demands change risk?}/{Matched delay, repetition, or delegation}/{Completion, failure, and recovery}}{
 \node[bx=smB] (q) at (\x,-1.45) {\q};
 \node[bx=smA] (c) at (\x,-3.05) {\c};
 \node[bx=smC] (m) at (\x,-4.65) {\m};
 \draw[ar] (q) -- (c); \draw[ar] (c) -- (m);}
\end{tikzpicture}}
\caption{Research roadmap linking five unresolved questions to comparisons that could settle them and observable outcomes.}
\label{fig:researchmap}
\Description{Five research priorities map propagation, memory, verification, recovery, and horizon questions to experimental comparisons and measured outcomes.}
\end{figure}

Each priority addresses a caveat in Section~\ref{sec:findings} or a gap in Section~\ref{sec:results} and names a comparison that would settle it (Figure~\ref{fig:researchmap}).

\textbf{OP1: Supported propagation graphs.} Exact step-level attribution reaches 14.2\% on Who\&When and 24.1\% on LongRCA~\citep{zhang2025whowhen,zhang2026longrca}. Dependency-guided attribution already improves on chronology-based judges~\citep{deng2026defa}, but attribution benchmarks contain surface cues that simple predictors exploit~\citep{xu2026auditattribution}, so comparisons should use held-out tasks, keep natural and injected failures separate, and report cue-control tests.

\textbf{OP2: Semantic memory and compute state under constrained resources.} Exact execution, approximate caches, summarization, learned compression, and external memory should be compared on fixed tasks at equal memory and latency, measuring decision fidelity and retraction. Injecting weight, cache, or computation faults and tracing them through decisions would close the R1 gap.

\textbf{OP3: Verifier portfolios with measurable blind spots.} Seeded violations of action, evidence, progress, and outcome requirements would let tests, policy monitors, guards, judges, fresh retrieval, and humans be compared on common outputs, with source sharing varied separately from model identity so that its effect on joint misses, false rejection, and cost can be measured.

\textbf{OP4: Recovery across uncertain commitment.} Benchmarks should inject timeouts before and after commitment, re-synthesized retries, and compensation failures, so that retry, checkpointing, staging, and reconciliation can be compared by post-recovery consistency and residual impact.

\textbf{OP5: Horizon interventions with valid stopping rules.} Varying dependency delay, repetition, or delegation on matched tasks with fixed harnesses would identify which structural demands change risk, rather than a step limit.

\subsection{Implications for research and practice}
\label{sec:implications}

Table~\ref{tab:practice} reverses Table~\ref{tab:controlcontracts}: it starts from what a deployment does and lists the controls and evidence to demand. Benchmark authors should publish evaluator versions, score populations, and horizon profiles; reviewers should separate model gains from changes in harness, budget, or task selection.

\begin{table}[!htbp]
\centering\footnotesize
\setlength{\tabcolsep}{4pt}
\caption{Practitioner guidance by deployment property; an agent with several properties needs each row and the composition obligations of Section~\ref{sec:mech7}.}
\label{tab:practice}
\begin{tabularx}{\linewidth}{@{}P{.22\linewidth}P{.36\linewidth}X@{}}
\toprule\textbf{If the agent\ldots}&\textbf{Controls to consider}&\textbf{Evidence to demand}\\\midrule
Releases irreversible effects&Staging; effect logs with replay-or-fork; provider deduplication&Duplicate and residual-impact rates under injected timeouts and restores\\
Reads untrusted content or tools&Data--authority separation; deterministic privilege policies; tool provenance&Attack success and utility under matched permissions\\
Keeps long-lived or shared memory&Provenance and consumers per item; invalidation; decision-based compression&Retraction success and decision fidelity at matched budgets\\
Relies on model judges&Independent evidence; seeded-violation evaluation&False acceptance on the worker's own errors\\
Coordinates several agents&Enforced ownership; versioned handoffs; bounded termination&Performance under shared-source faults at equal budget\\
Runs for hours or sessions&Task state verified against the environment; reconciliation&$\mathrm{pass}^k$ and detection latency by horizon component\\
Requires human approval&Review packets of effects and evidence; policy set for a reliability target&Review load and residual errors at that target\\
\bottomrule
\end{tabularx}

\end{table}

\end{document}